\PassOptionsToPackage{nopatch=footnote,expansion=false}{microtype}
\documentclass[accepted]{uai2026} 

\usepackage{amsmath,amsfonts,bm}

\def\eqref#1{equation~\ref{#1}}

\def\1{\bm{1}}

\DeclareMathAlphabet{\mathsfit}{\encodingdefault}{\sfdefault}{m}{sl}
\SetMathAlphabet{\mathsfit}{bold}{\encodingdefault}{\sfdefault}{bx}{n}

\usepackage{xspace}

\usepackage[american]{babel}
\usepackage{natbib}
\usepackage{hyperref}
\usepackage{tabularx}
\usepackage{makecell}
\usepackage{booktabs}
\usepackage{mathtools}
\usepackage{booktabs}
\usepackage{tikz} 
\usepackage{graphicx} 
\usepackage{float}
\usepackage{algorithm}
\usepackage{algpseudocode}
\usepackage{algorithmicx}
\usepackage{amsmath}
\usepackage{placeins} 
\usepackage{multirow}
\usepackage{amssymb}
\usepackage{bm}
\usepackage{dsfont}
\usepackage{url}
\usepackage{float} 
\usepackage{longtable}
\usepackage{tabularx}
\usepackage{caption}
\usepackage{subcaption}
\usepackage{siunitx}
\definecolor{citecol}{RGB}{139,0,0}
\definecolor{linkcol}{RGB}{244, 187, 68}
\usepackage{amsfonts}       
\usepackage{nicefrac}       
\usepackage{comment}
\usepackage{amsmath}   
\usepackage{pifont}      
\usepackage{tablefootnote} 
\usepackage[table]{xcolor}
\usepackage{environ}
\usepackage{enumitem}
\usepackage{subcaption}
\usepackage{wrapfig}  
\usepackage{adjustbox}
\usepackage{setspace}     
\usepackage{array} 
\usepackage{makecell}
\usepackage{arydshln}
\usepackage{listings}
\usepackage{pdfpages}
\usepackage{courier}
\usepackage{colortbl}  
\definecolor{gray94}{gray}{.94}
\definecolor{gray90}{gray}{.90}

\definecolor{darkgreen}{RGB}{0,150,0}
\definecolor{darkred}{RGB}{200,0,0}

\usepackage{color-edits}
\usepackage{dsfont}
\usepackage{bbding}
\usepackage[most]{tcolorbox} %
\usepackage[utf8]{inputenc}  %
\definecolor{darkpastelgreen}{rgb}{0.01, 0.75, 0.24}

\UseRawInputEncoding
\usepackage{fontawesome}
\usepackage{tikz}
\usetikzlibrary{arrows.meta,calc,positioning}

\tcbuselibrary{listings, skins}
\tcbset{
  commonstyle/.style={
    breakable,         
    enhanced,          
    boxrule=1pt,       
    fonttitle=\bfseries,
    colbacktitle=white,  
  }
}

\title{Interpretable Causal Discovery via Causal-Effect Constraints}

\author[1]{Cixuan Zhang}
\author[2]{Guy Van den Broeck}
\author[2]{Benjie Wang}

\affil[1]{%
    Computer Science Dept.\\
    Yale University\\
    New Haven, Connecticut, USA
}
\affil[2]{%
    Computer Science Dept.\\
    University of California, Los Angeles\\
    Los Angeles, California, USA
}

\begin{document}

\maketitle

\footnotetext{Our code is available at \url{https://github.com/ZCX031116/MLS-Framework}.}

\begin{abstract}

Causal discovery aims to uncover the underlying causal relationships given data generated from a system. The goal, however, is not merely to predict causal edges given data, but also to be able to interpret and explain either observed or hypothesized phenomena, such as a particularly large causal effect. We consider this task of \emph{conditional} causal discovery and cast it as a Bayesian inference problem, in which we target the posterior over causal graphs and parameters conditional on an event such as a causal-effect constraint. Unfortunately, this poses a computational challenge: existing approaches to Bayesian causal discovery struggle when the event has small posterior mass. To address this, we adapt rare-event estimation techniques to perform inference the joint graph-parameter space. Our method gradually drives a particle population toward the constrained region while maintaining samples that approximate the conditional posterior. Empirical evaluation on synthetic graphs validates the accuracy of our approach at small and large scales, and we show in a case study on the Sachs protein dataset how our method can be used to aid scientific exploration by providing pathway-level summaries.

\end{abstract}

\section{Introduction}\label{sec:motiv}
Causal discovery is often motivated not only by prediction, but by the need to obtain interpretable and actionable explanations of how a system works. In scientific domains, a practitioner may not simply ask which graph is most probable given the data; rather, they may ask which causal mechanisms could explain a particular domain-relevant phenomenon. For example, in the well-known Sachs study of protein interactions \citep{sachs2005causal}, one may want to understand which directed pathways could support an unusually large effect from one protein to another, and whether such pathways suggest plausible interventions or follow-up experiments. 
%



\begin{figure*}[t]
\centering
\resizebox{\textwidth}{!}{
\begin{tikzpicture}[
    >=Latex,
    font=\small,
    panel/.style={
        draw=black!70,
        rounded corners=4pt,
        fill=gray!4,
        minimum width=5.35cm,
        minimum height=4.75cm,
        inner sep=0pt
    },
    title/.style={
        font=\bfseries\large,
        align=center
    },
    vnode/.style={
        circle,
        draw=black!75,
        fill=white,
        minimum size=20pt,
        inner sep=0pt,
        font=\scriptsize
    },
    ghostnode/.style={
        circle,
        draw=black!25,
        fill=white,
        text=black!35,
        minimum size=20pt,
        inner sep=0pt,
        font=\scriptsize
    },
    hnode/.style={
        circle,
        draw=orange!75!black,
        fill=orange!18,
        minimum size=20pt,
        inner sep=0pt,
        font=\scriptsize
    },
    edge/.style={
        -Latex,
        line width=0.55pt,
        draw=black!35
    },
    ghostedge/.style={
        -Latex,
        line width=0.55pt,
        draw=black!18
    },
    hedge/.style={
        -Latex,
        line width=1.30pt,
        draw=orange!85!black
    },
    flow/.style={
        -Latex,
        line width=1.0pt,
        draw=black!70
    },
    arrowlabel/.style={
        font=\scriptsize,
        align=center,
        inner sep=0pt,
        text=black!85
    },
    foot/.style={
        font=\scriptsize,
        align=center,
        text=black!75
    }
]

\node[panel] (P) at (0,0) {};
\node[title] at ($(P.north)+(0,-0.38)$) {Posterior samples};

\node[foot] at ($(P.north)+(0,-0.86)$)
    {$p(G,B\mid\mathcal D)$: many plausible DAGs};

\node[vnode] (p_i) at ($(P.center)+(-1.35,0.35)$) {$i$};
\node[vnode] (p_a) at ($(P.center)+(-0.35,0.72)$) {};
\node[vnode] (p_b) at ($(P.center)+(-0.35,-0.02)$) {};
\node[vnode] (p_j) at ($(P.center)+(0.75,0.35)$) {$j$};

\draw[edge] (p_i) -- (p_a);
\draw[edge] (p_i) -- (p_b);
\draw[edge] (p_a) -- (p_j);
\draw[edge] (p_b) -- (p_j);

\node[ghostnode] (pg_i) at ($(P.center)+(-0.95,-1.10)$) {$i$};
\node[ghostnode] (pg_a) at ($(P.center)+(-0.05,-0.9)$) {};
\node[ghostnode] (pg_b) at ($(P.center)+(-0.05,-1.90)$) {};
\node[ghostnode] (pg_j) at ($(P.center)+(0.88,-1.10)$) {$j$};

\draw[ghostedge] (pg_i) -- (pg_a);
\draw[ghostedge] (pg_a) -- (pg_b);
\draw[ghostedge] (pg_b) -- (pg_j);

\node[panel] (E) at (6.65,0) {};
\node[title] at ($(E.north)+(0,-0.38)$)
    {Causal-effect constraint};

\node[align=center, font=\footnotesize] at ($(E.center)+(0,0.8)$)
    {$\mathcal E_{ij}^{+}(t)=\{(G, B):\mathrm{CE}_{ij}(G, B)\ge t\}$};

\node[align=center, font=\footnotesize, text=black!55]
    at ($(E.center)+(0,0.50)$) {};

\draw[black!35, line width=0.65pt]
    ($(E.center)+(-1.30,-0.78)$)
    .. controls ($(E.center)+(-0.88,-0.20)$)
           and ($(E.center)+(-0.24,-0.22)$)
        .. ($(E.center)+(0.14,-0.74)$)
    .. controls ($(E.center)+(0.50,-1.00)$)
           and ($(E.center)+(0.98,-0.90)$)
        .. ($(E.center)+(1.28,-0.82)$);

\draw[dashed, line width=0.65pt, draw=black!65]
    ($(E.center)+(0.62,-1.10)$) --
    ($(E.center)+(0.62,-0.36)$);

\node[font=\scriptsize] at ($(E.center)+(0.82,-0.38)$) {$t$};

\node[font=\footnotesize] at ($(E.south)+(0,0.30)$)
    {rare posterior event};

\node[panel] (C) at (13.30,0) {};
\node[title] at ($(C.north)+(0,-0.38)$)
    {Conditional posterior};

\node[foot] at ($(C.north)+(0,-0.95)$)
    {$p(G,B\mid\mathcal D,\mathcal E_{ij}(t))$};

\node[hnode] (c_i) at ($(C.center)+(-1.3,0.03)$) {$i$};
\node[hnode] (c_a) at ($(C.center)+(-0.25,0.70)$) {};
\node[vnode] (c_b) at ($(C.center)+(-0.25,-0.6)$) {};
\node[hnode] (c_j) at ($(C.center)+(0.9,0.03)$) {$j$};

\draw[edge] (c_i) -- (c_b);
\draw[edge] (c_b) -- (c_a);
\draw[edge] (c_a) -- (c_j);
\draw[edge] (c_b) -- (c_j);
\draw[edge] (c_i) -- (c_a);

\draw[hedge] (c_i) -- (c_b);
\draw[hedge] (c_b) -- (c_a);
\draw[hedge] (c_a) -- (c_j);

\node[foot] at ($(C.south)+(0,0.60)$)
    {high-frequency pathways\\explain the event};

\draw[flow] (P.east) -- (E.west)
    node[midway, above=3pt, arrowlabel]
    {score by\\$\mathrm{CE}_{ij}$};

\draw[flow] (E.east) -- (C.west)
    node[midway, above=3pt, arrowlabel]
    {MLS\\+ MCMC};

\end{tikzpicture}
}

\caption{Overview of conditional causal discovery. Starting from the
unconstrained Bayesian posterior over causal graphs and edge weights, we
condition on a user-specified extreme causal-effect event and obtain a
constrained posterior whose samples can be summarized at the pathway level.}
\label{fig:ccd_overview}
\end{figure*}
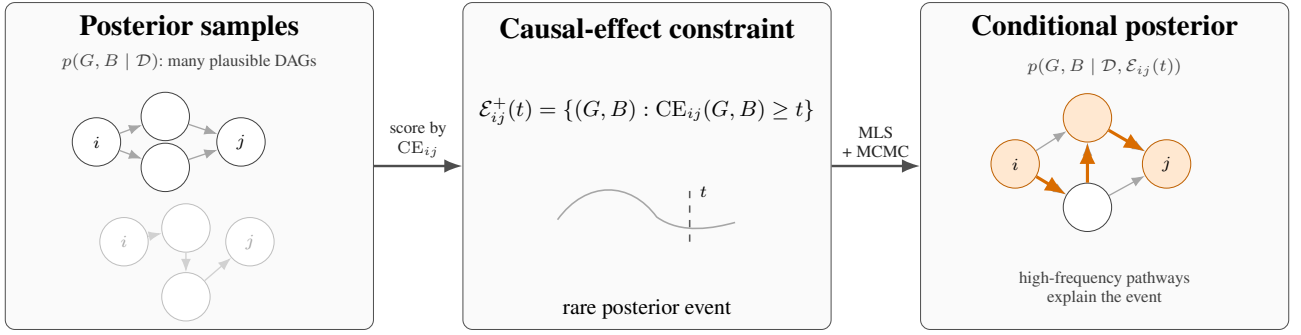

In this work, we formulate this type of ``what-if'' analysis as a \emph{conditional} form of causal discovery. Given observational data, we seek causal structures that are both statistically plausible and consistent with a user-specified constraint, such as a large causal effect from node $i$ to node $j$.
This presupposes that there could be many graphs that explain the data well, and as such requires a treatment of epistemic uncertainty. 
We take a Bayesian approach to this problem in which we represent uncertainty over graph structure and parameters. Rather than committing to a single estimated graph, we consider the posterior distribution over directed acyclic graphs ($G$) and parameters ($\theta$). Our target is the posterior conditioned on an event determined by the graph and parameters, such as the event that the signed causal effect from  $i$ to $j$ exceeds a threshold $t$. 
The goal is to characterize the causal structures and directed pathways that remain plausible under the causal-effect constraint, while also estimating the posterior probability of the event itself. 

Conditional causal discovery poses a computational challenge that is not well handled by current tools. In particular, the event of interest may be rare under the unconstrained posterior, especially when the causal effect threshold $t$ is large, or when we impose a combination of constraints. Standard posterior samplers based on MCMC may produce few or no samples satisfying the constraint, and simple rejection-based conditioning becomes inefficient. Even when constraint-satisfying samples are obtained, poor mixing can lead to unreliable statistical summaries of e.g. different causal paths. 

We therefore cast conditional causal discovery under strong (e.g. extreme-effect) constraints as a rare-event posterior inference problem. Our approach integrates adaptive multilevel splitting (AMS) \citep{cerou2007adaptive} with an MCMC kernel over the joint graph--parameter space. Starting from posterior samples obtained from any Bayesian causal discovery method, our method progressively tightens the effect threshold while maintaining a representative particle population, producing both tail-probability estimates and conditional posterior samples. These samples can then be used to summarize which edges and directed pathways are most characteristic of the extreme-effect regime.

Our contributions are as follows:
\begin{itemize}[leftmargin=*]
\item We formulate conditional causal discovery as posterior inference under extreme-effect constraints, targeting both constrained posterior samples and posterior tail probabilities.
\item We provide a practical rare-event inference procedure that can estimate small posterior tail probabilities while producing representative graph--parameter samples from the corresponding constrained posterior.
\item We validate the method on linear-Gaussian benchmarks with ($d\in{4,8,16,32}$) and present a Sachs dataset case study showing how extreme-effect conditioning enables pathway-level interpretation and hypothesis generation.
\end{itemize}

\section{Related Work}\label{sec:works}

Causal discovery under background knowledge or structural constraints has been widely studied. Such constraints often encode qualitative statements about graph structure, including required or forbidden edges, paths, or ancestral relations \citep{meek1995causal,borboudakis2012incorporating,anand2023causal}. Prior work has considered both identifiability and algorithmic procedures for learning graphs subject to these constraints \citep{chen2016learning,chen2024identifying}. A related line of work uses interventional data as an additional source of information beyond pure observations \citep{hauser2015jointly,brouillard2020differentiable}. Instead, we are primarily motivated by interpretability rather than encoding fixed knowledge. Our technical approach differs in that the constraint is encoded flexibly as a quantitative score function, and in particular we encode constraints corresponding to the value of a causal-effect functional rather than a local structural statement. Moreover, rather than seeking exact identification of a single graph, we target a conditional Bayesian posterior that retains uncertainty over both graphical and parametric structure.

Another line of work combines causal discovery with extreme value theory to
infer causal directions when causal mechanisms are most visible in the tails of
the observed distribution \citep{gnecco2021causal,pasche2023causal,bodik2024causality}.
In these approaches, extremes are primarily a feature of the data-generating
distribution and are used for identifiability or structure recovery. Our
motivation is different: in our setting, the extreme event is a user-specified
constraint on a causal-effect functional, and the inferential target is the
posterior distribution conditioned on that constraint. Thus, our goal is not to
identify causal directions from heavy-tailed observations, but to characterize
which posterior graphs, parameters, and pathways explain an unusually large or
small causal effect.

For posterior inference over causal graphs, a common approach is to use Markov chain Monte Carlo (MCMC) sampling over DAGs or higher-level representations such as orders \citep{friedman2003being,kuipers2017partition,viinikka2020towards,giudice2023bayesian}. More recent work has developed efficient DAG-space MCMC samplers with locally informed and adaptive proposals, including PARNI-DAG \citep{liang2023structure}, which constructs adaptive random neighborhoods guided by posterior information and can exploit a pre-tuned skeleton to improve scalability. These graph posterior samples are useful not only for representing structural uncertainty, but also for downstream causal inference; recent benchmarking work evaluates Bayesian causal discovery methods through downstream treatment-effect estimation \citep{emezue2023benchmarking}.

Alternative approaches approximate the posterior via variational inference over graphs \citep{annadani2021variational,lorch2021dibs,cundy2021bcd,wang2022tractable,deleu2022bayesian,rittel2023specifying,toth2024effective}. These methods aim either to generate graph samples reflecting structural uncertainty or to support Bayesian model averaging for causal inference \citep{toth2022active}. 

Our computational approach depends on the rare-event simulation literature, particularly adaptive multilevel splitting \citep{cerou2007adaptive}. Splitting methods estimate small probabilities by introducing intermediate thresholds and repeatedly propagating a particle population toward rarer events. We adapt this idea to Bayesian causal discovery by defining levels through a causal-effect score and using MCMC moves over the joint graph--parameter space. This produces both an estimate of the posterior probability of the extreme-effect event and samples from the corresponding conditional posterior.

\section{Preliminaries}\label{sec:theory}
\paragraph{Causal Bayesian Networks}
A Bayesian network (BN) $(G,\theta)$ is a probabilistic model $p(\bm{X})$
over $d$ variables $\bm{X}={X_1,\ldots,X_d}$, specified by a directed
acyclic graph (DAG) $G$ and mechanism parameters $\theta$. The graph encodes
conditional independencies, while $\theta_i$ parameterizes the conditional
distribution of $X_i$ given its parents. The joint distribution factorizes as
\[
p(\bm{X}\mid G,\theta)
=
\prod_{i=1}^d
p\left(X_i \mid \mathrm{pa}_G(X_i), \theta_i\right),
\]
where $\mathrm{pa}_G(X_i)$ denotes the parents of $X_i$ in $G$.

In this paper, the generic mechanism parameters $\theta$ are instantiated by a
linear-Gaussian structural equation model. Specifically, the observed variables
satisfy $\bm{X}=\bm{X}B+\bm{\epsilon}$,
where $B\in\mathbb{R}^{d\times d}$ is the weighted adjacency matrix and
$\bm{\epsilon}\sim \mathcal N(\bm b,\Sigma)$, with
$\bm b\in\mathbb{R}^d$ and diagonal
$\Sigma\in\mathbb{R}_{\ge 0}^{d\times d}$. For a given DAG $G$, we impose
$B_{ij}=0$ whenever $i$ is not a parent of $j$ in $G$.

Causal Bayesian networks \citep{spirtes2000causation,Pearl09Causality}
add a causal interpretation to the directed edges in $G$: they describe how
the joint distribution changes under interventions. In the linear-Gaussian SEM,
the signed total causal effect from $X_i$ to $X_j$ is the derivative of the
post-intervention mean of $X_j$ with respect to an intervention on $X_i$. It
has the closed form
\begin{equation}
\label{eq:linear_sem_total_effect}
\mathrm{CE}_{ij}(G,B)
\triangleq
\mathrm{CE}(i\to j\mid G,B)
=
\left[(I-B)^{-1}\right]_{ij}
\end{equation}
following the standard total-effect formula for linear structural equation
models \citep{sobel1990effect}. Since $G$ is acyclic, $B$ is nilpotent after
a topological ordering, so $(I-B)^{-1}=I+B+B^2+\cdots+B^{d-1}$.
Thus, the $(i,j)$ entry of $(I-B)^{-1}$ aggregates the products of edge
weights along all directed paths from $X_i$ to $X_j$. This path-sum
interpretation is useful below because our conditional posterior summaries focus
on which edges and directed pathways explain unusually large or small total
effects.

Throughout the rest of the paper, we write a graph--weight state as $Z=(G,B)$,
and abbreviate the signed causal effect as $\mathrm{CE}_{ij}(Z)=\mathrm{CE}_{ij}(G,B)$.
For a target ordered pair $(i,j)$ and threshold $t>0$, we use the phrase
``extreme causal effect'' to mean that the total causal effect falls in a
user-specified posterior tail region. In particular, we define the right- and
left-tail events
\begin{equation}
\label{eq:extreme_effect_events}
\begin{aligned}
\mathcal{E}_{ij}^{+}(t)
&=
\left\{Z: \mathrm{CE}_{ij}(Z) \ge t\right\}, \\
\mathcal{E}_{ij}^{-}(t)
&=
\left\{Z: \mathrm{CE}_{ij}(Z) \le -t\right\}.
\end{aligned}
\end{equation}
When the sign is clear from context, we write $\mathcal{E}_{ij}(t)$ for either
tail event. These events play two roles in our inference problem. First, we
estimate their posterior probability, such as
$\mathbb{P}(\mathcal{E}_{ij}^{\pm}(t)\mid\mathcal D)$. Second, we use them to
define the constrained posterior $p\left(G,B\mid \mathcal D,\mathcal{E}_{ij}^{\pm}(t)\right)$, from which we draw graph--weight samples for pathway-level summaries.

\paragraph{Bayesian Causal Discovery}
\textit{Causal discovery} \citep{koller2009probabilistic,glymour2019review}
is the problem of inferring the DAG $G$ responsible for generating an observed
dataset $\mathcal D$. We make the common assumption of causal sufficiency,
meaning that there are no latent confounders. Even under this assumption, a
single DAG may not be reliably identifiable from finite observational data due
to sampling uncertainty and Markov equivalence. Bayesian causal discovery
therefore represents uncertainty through a posterior distribution over graphs
and parameters rather than committing to a single structure.

We place a user-specified prior $p(G)$ over DAGs and use the BGe marginal
likelihood $p(\mathcal D\mid G)$ for linear-Gaussian models
\citep{geiger1994learning,geiger2002parameter}. The method itself does not
require a particular graph prior; in the experiments, we use a sparse
Erd\H{o}s--R'enyi DAG prior, with the exact sparsity settings reported in
Appendix~\ref{sec:exp_setting}. Given $G$, the posterior over edge weights
factorizes by node:
\[
p(B\mid G,\mathcal D)
=
\prod_{j=1}^d
p\left(B_{\mathrm{pa}_G(X_j),j}\mid G,\mathcal D\right),
\]
where coefficients outside the parent set of $X_j$ are fixed to zero and each
nonzero incoming-coefficient block follows a multivariate $t$-distribution
\citep{viinikka2020towards}. This node-wise factorization is used later by the
MCMC mutation kernel: when a graph proposal changes a node's parent set, only
the affected incoming-coefficient blocks need to be refreshed from their
conditional posterior.

The joint posterior over graph--weight states is
\begin{equation}
\label{eq:joint_posterior}
p(G,B\mid \mathcal D)
\propto
p(G),p(\mathcal D\mid G),p(B\mid G,\mathcal D).
\end{equation}
We write $\pi(Z)=\pi(G,B)=p(G,B\mid \mathcal D)$ for this unconstrained posterior in the method section for compactness.

\section{Conditional Causal Discovery}\label{sec:method}

In this section, we formulate conditional causal discovery as posterior inference under user-specified causal-effect constraints, and then describe how we sample from the resulting conditional posterior. The main challenge is that the constraint may have low probability under the posterior. If ordinary posterior samples almost never satisfy the event, then rejection sampling gives both unstable probability estimates and too few constrained samples for pathway-level interpretation. Our solution is to convert each query into a scalar score $h(Z)$, where larger values indicate greater progress toward the desired event, and then use adaptive multilevel splitting to reach the event through a sequence of easier conditional problems.

\subsection{Conditional causal discovery as a score-level problem}
\label{ssec:conditional_score_problem}

Let $Z=(G,B)$ denote a graph--weight state and let $\pi(Z)=p(G,B\mid \mathcal D)$ be the unconstrained posterior from Eq.~\eqref{eq:joint_posterior}. Throughout this section, we use $\pi(\mathcal A)$ to denote the posterior mass of an event $\mathcal A$. A conditional query is specified by a score function $h:\mathcal Z\to\mathbb R$ and a final level $\lambda_\star$. The score-level event, its posterior probability, and the corresponding constrained posterior are
\begin{equation}
\label{eq:score_level_target}
\begin{aligned}
    \mathcal A_{\lambda_\star}
    &=
    \{Z:h(Z)\ge \lambda_\star\},\\
    p_\star
    &=
    \pi(\mathcal A_{\lambda_\star}),\\
    \pi_\star(Z)
    &=
    \pi\!\left(Z\mid\mathcal A_{\lambda_\star}\right).
\end{aligned}
\end{equation}
This formulation separates the scientific query from the computational procedure: once the query has been written through $h$ and $\lambda_\star$, the same sampler can be applied to single effects, left-tail effects, or multiple simultaneous constraints.

For a single ordered pair $(i,j)$ and threshold $t>0$, the right-tail event $\mathcal E_{ij}^{+}(t)$ is represented by choosing $h(Z)=\mathrm{CE}_{ij}(Z)$ and $\lambda_\star=t$. The left-tail event $\mathcal E_{ij}^{-}(t)$ is represented by choosing $h(Z)=-\mathrm{CE}_{ij}(Z)$ and $\lambda_\star=t$. Thus both tails are written as the same score-level event $\mathcal A_{\lambda_\star}=\{Z:h(Z)\ge\lambda_\star\}$, which avoids requiring separate algorithms for positive and negative extreme effects.

We also support conjunctions of multiple causal-effect constraints. Let $\mathcal C=\{(i_c,j_c,\bowtie_c,t_c,\kappa_c)\}_{c=1}^{C}$ be a collection of $C$ constraints, where $\bowtie_c\in\{\ge,\le\}$ gives the inequality direction, $t_c\in\mathbb R$ is the threshold, and $\kappa_c>0$ is an optional scale factor. Define $\delta_c=+1$ when $\bowtie_c=\ge$ and $\delta_c=-1$ when $\bowtie_c=\le$. The normalized margin of constraint $c$ and the aggregate score are
\begin{equation}
\label{eq:multi_constraint_score}
\begin{aligned}
    m_c(Z)
    &=
    \frac{\delta_c\bigl(\mathrm{CE}_{i_cj_c}(Z)-t_c\bigr)}{\kappa_c},\\
    h_{\mathcal C}(Z)
    &=
    \min_{1\le c\le C}m_c(Z).
\end{aligned}
\end{equation}
The margin $m_c(Z)$ is nonnegative exactly when the $c$-th constraint is satisfied. Therefore $h_{\mathcal C}(Z)\ge0$ means that all constraints are satisfied, so the joint event is $\mathcal E_{\mathcal C}=\{Z:h_{\mathcal C}(Z)\ge0\}$. Interval constraints can be represented by including both a lower-bound and an upper-bound inequality.

\subsection{Adaptive multilevel splitting for rare posterior events}
\label{ssec:amls_background}

Extreme-effect events can have very small posterior probability. A direct posterior sampler with $M$ samples will produce only about $Mp_\star$ constrained samples on average, which may be close to zero if $p_\star$ is small. Adaptive multilevel splitting addresses this by replacing one difficult rare-event problem with a sequence of easier conditional problems. Intuitively, particles are first asked to reach a moderately high score level, then a higher one, and so on until they reach the target level $\lambda_\star$. At each stage, particles that have made sufficient progress are retained and resampled, while MCMC mutation restores diversity within the current truncated posterior.

Let $\lambda_0<\lambda_1<\cdots<\lambda_K=\lambda_\star$ be increasing score levels, with $\lambda_0=-\infty$ corresponding to the unconstrained posterior. Define $\mathcal A_k=\{Z:h(Z)\ge\lambda_k\}$. Because these events are nested, the rare-event probability decomposes as
\begin{equation}
\label{eq:splitting_product}
    p_\star
    =
    \pi(\mathcal A_K)
    =
    \prod_{k=0}^{K-1}
    \pi(\mathcal A_{k+1}\mid \mathcal A_k).
\end{equation}
The advantage is that each factor in this product can be much larger than $p_\star$ itself, making it estimable with a moderate number of particles.

We use the adaptive version of multilevel splitting \citep{kahn1951estimation,guyader2011simulation,cerou2007adaptive}, which chooses intermediate levels from the particle population rather than requiring them to be fixed in advance. At level $k$, the particle population $\mathcal P_k=\{Z_{k,n}\}_{n=1}^{N}$ is first mutated so that it approximately follows the level-truncated posterior
\begin{equation}
\label{eq:level_truncated_posterior}
    \pi_{\lambda_k}(Z)
    \propto
    \pi(Z)\mathbf 1\{h(Z)\ge\lambda_k\}.
\end{equation}
We then evaluate the scores $h(Z_{k,n})$ and choose the next level as an empirical quantile:
\begin{equation}
\label{eq:adaptive_level}
\begin{aligned}
    \widetilde\lambda_{k+1}
    &=
    Q_{1-\rho}\!\left(\{h(Z_{k,n})\}_{n=1}^{N}\right),\\
    \lambda_{k+1}
    &=
    \min\{\widetilde\lambda_{k+1},\lambda_\star\}.
\end{aligned}
\end{equation}
Here $\rho\in(0,1)$ is the survival fraction and $Q_{1-\rho}$ is the empirical $(1-\rho)$-quantile. The estimated conditional factor at this level is the fraction of particles that survive the new threshold,
\begin{equation}
\label{eq:survival_factor}
    \widehat\beta_{k+1}
    =
    \frac{1}{N}
    \sum_{n=1}^{N}
    \mathbf 1\{h(Z_{k,n})\ge\lambda_{k+1}\}.
\end{equation}
The survivors are resampled with replacement to form the starting population for the next level. The final probability estimate is $\widehat p_\star=\prod_{k=0}^{K-1}\widehat\beta_{k+1}$, and after the final mutation step the particles approximate samples from $\pi_\star(Z)$.

\begin{algorithm}[t]
\caption{Adaptive multilevel splitting for conditional causal discovery}
\label{alg:amls_compact}
\footnotesize
\begin{algorithmic}[1]
\Require Posterior $\pi(Z)$, score $h$, target level $\lambda_\star$,
initial particles $\mathcal P$, particle size $N$, survival fraction $\rho$,
mutation steps $m$, maximum levels $K_{\max}$.
\Ensure Tail-probability estimate $\widehat p_\star$ and constrained particles $\mathcal P$.

\State Initialize $\mathcal P=\{Z_n\}_{n=1}^{N}$ using an approximate posterior sampler or warm-start routine targeting $\pi(Z)$.
\State Set $\lambda_0\gets -\infty$ and $\widehat p_\star\gets 1$.

\For{$k=0,\ldots,K_{\max}-1$}
    \State $\mathcal P\gets \mathrm{Mutate}_{\lambda_k}(\mathcal P;m)$.
    \State Compute scores $s_n\gets h(Z_n)$ for all $Z_n\in\mathcal P$.
    \State Set
    $\lambda_{k+1}\gets
    \min\{Q_{1-\rho}(s_1,\ldots,s_N),\lambda_\star\}$.
    \State Let
    $\mathcal S\gets
    \{Z_n\in\mathcal P:s_n\ge \lambda_{k+1}\}$.
    \State Estimate
    $\widehat\beta_{k+1}\gets |\mathcal S|/N$ and update
    $\widehat p_\star\gets \widehat p_\star\widehat\beta_{k+1}$.

    \If{$|\mathcal S|=0$}
        \State \Return $0,\emptyset$.
    \EndIf

    \State Resample $N$ particles from $\mathcal S$ with replacement to form $\mathcal P$.

    \If{$\lambda_{k+1}=\lambda_\star$}
        \State $\mathcal P\gets \mathrm{Mutate}_{\lambda_\star}(\mathcal P;m)$.
        \State \Return $\widehat p_\star,\mathcal P$.
    \EndIf
\EndFor

\State \Return $\widehat p_\star,\mathcal P$.
\Statex
\Statex \textit{Here $\mathrm{Mutate}_{\lambda}(\mathcal P;m)$ denotes $m$ Metropolis--Hastings}
\Statex \textit{steps per particle targeting
$\pi_{\lambda}(Z)\propto \pi(Z)\mathbf 1\{h(Z)\ge \lambda\}$.}
\end{algorithmic}
\end{algorithm}

\subsection{Particle initialization}
\label{ssec:particle_initialization}

The splitting procedure should be viewed as a rare-event wrapper around Bayesian causal discovery rather than as a replacement for it. In principle, the initial population can be obtained from any posterior sampler that approximately targets $\pi(Z)$, including a long MCMC chain, an informed DAG-space sampler, or another Bayesian structure-learning method. The role of multilevel splitting is then to take these ordinary posterior samples and concentrate computation on the score-level region $\mathcal A_{\lambda_\star}$.

In our implementation, we use a simple warm-start scheme. We first draw sparse DAGs $G_{0,n}$ from the structural prior $p(G)$, using an Erd\H{o}s--R\'enyi DAG prior in the experiments with edge inclusion probability $p_{\mathrm{edge}}$ chosen to match a target expected number of edges per node. Given each initial graph, we draw weights from the conjugate posterior $B_{0,n}\sim p(B\mid G_{0,n},\mathcal D)$ and set $Z_{0,n}=(G_{0,n},B_{0,n})$.

Because these raw initial particles are not assumed to be exact samples from $\pi(Z)$, we use the inner MCMC kernel at the initial level $\lambda_0=-\infty$ as a posterior warm-up before adapting the first nontrivial threshold. This makes the first population used in the splitting product approximate the unconstrained posterior, while retaining a simple and scalable initialization procedure.

\subsection{Inner Metropolis--Hastings over joint \texorpdfstring{$(G,B)$}{(G,B)}}
\label{ssec:mcmc}

At each score level $\lambda_k$, mutation must preserve the level-truncated target $\pi_{\lambda_k}(Z)$ from Eq.~\eqref{eq:level_truncated_posterior}. We use a blocked Metropolis--Hastings kernel over the joint state $Z=(G,B)$, alternating between graph updates and coefficient updates. This is important because the rare-event constraint depends on both structure and weights: changing only $G$ can leave the sampler stuck at a fixed set of coefficients, while changing only $B$ cannot explore alternative pathways.

With probability $p_{\mathrm{struct}}$, we propose a structure move. The new graph $G'$ is drawn from a proposal $q_G(\cdot\mid G)$, instantiated either as Structure-MCMC, which adds, deletes, or reverses a single edge while rejecting cyclic graphs, or as PARNI-DAG \citep{liang2023structure}, which uses locally informed adaptive neighborhoods guided by posterior edge information. After proposing $G'$, we refresh only the coefficient blocks whose parent sets changed. For each affected node $v$, we sample
\begin{equation}
\label{eq:structure_weight_refresh}
    B'_{\mathrm{pa}_{G'}(v),v}
    \sim
    p\!\left(B_{\mathrm{pa}_{G'}(v),v}\mid G',\mathcal D\right),
\end{equation}
set coefficients for absent edges to zero, and copy all unchanged blocks from $B$.

With probability $1-p_{\mathrm{struct}}$, we propose a weight move at fixed structure. We keep $G'=G$, select a node $v$ with at least one parent, and resample its incoming coefficient block,
\begin{equation}
\label{eq:weight_refresh}
    B'_{\mathrm{pa}_{G}(v),v}
    \sim
    p\!\left(B_{\mathrm{pa}_{G}(v),v}\mid G,\mathcal D\right),
\end{equation}
leaving all other entries unchanged. These blocked refreshes are inexpensive under the conjugate linear-Gaussian model because each conditional coefficient posterior is a multivariate $t$ distribution.

Let $q(Z'\mid Z)$ denote the complete proposal density, including the selected move type, the graph proposal when applicable, and the coefficient-refresh density. Since the current state already satisfies $h(Z)\ge\lambda_k$, the level-$\lambda_k$ acceptance probability is
\begin{equation}
\label{eq:level_mh_acceptance}
\begin{aligned}
    a_{\lambda_k}(Z,Z')
    &=
    \mathbf 1\{h(Z')\ge\lambda_k\}\\
    &\quad\times
    \min\!\left\{
        1,
        \frac{\pi(Z')q(Z\mid Z')}{\pi(Z)q(Z'\mid Z)}
    \right\}.
\end{aligned}
\end{equation}
Thus, proposals that violate the current score-level constraint are rejected immediately, while feasible proposals are accepted according to the usual Metropolis--Hastings ratio for the unconstrained posterior and the proposal probabilities. Repeating these moves after each resampling step helps remove duplicate particles and produces a more representative approximation to the constrained posterior.
Zs

\section{Experiments}\label{sec:experiment}
\subsection{Experiment Setup}
\label{sec:exp_setup}

We evaluate the proposed conditional causal discovery procedure on synthetic
linear-Gaussian datasets generated from random Erd\H{o}s--R\'enyi DAGs with
$d\in\{4,8,16,32\}$. For each dimension, we first sample a data-generating DAG
$G^\star$ using an Erd\H{o}s--R\'enyi DAG generator with target sparsity
approximately $2d$ edges. The generator samples an acyclic ordering, draws only
order-compatible directed edges, and returns an adjacency matrix $G^\star$.
The exact edge-sampling rule, including the edge probability or edge budget used
for each $d$ and the small-graph handling for $d=4$, is provided in
Appendix~\ref{sec:exp_setting}.

Given $G^\star$, we draw raw edge weights independently as
$\widetilde B_{ij}\sim\mathcal N(0,1)$ and mask them by the sampled adjacency,
so that $B^\star=\widetilde B\odot G^\star$. Equivalently,
$B^\star_{ij}=\widetilde B_{ij}$ if $G^\star_{ij}=1$ and
$B^\star_{ij}=0$ otherwise. We then generate an observational dataset
$\mathcal D$ of size $n_{\mathrm{obs}}=1000$ from the linear-Gaussian SEM
specified by $(G^\star,B^\star)$.

We evaluate our multilevel splitting framework instantiated with either the
Structure-MCMC kernel or the PARNI-DAG kernel \citep{liang2023structure}. We
compare against four baselines: (i) exhaustive enumeration, which is a
gold-standard baseline feasible for $d=4$; (ii) DiBS \citep{lorch2021dibs};
(iii) OrderSPN \citep{wang2022tractable}; and (iv) long single-chain MCMC using
the same structural kernels as the multilevel splitting framework. Detailed
sampler hyperparameters, thresholds, and evaluation settings are provided in
Appendix~\ref{sec:exp_setting}.

The experiments are organized around three questions: small-graph accuracy,
multi-effect conditioning behavior, and scalability beyond the enumerable
setting. We describe each question in the corresponding subsection below.

\subsection{Correctness of Conditional Causal Discovery}
\label{sec:correctness}
The correctness experiments are designed to validate two complementary aspects
of the method. The single-effect experiment checks the numerical accuracy of
posterior tail-probability estimation against an exact enumerative reference. The
multi-effect experiment checks the conditioning operator itself: when the
conditioning event is constructed around ground-truth causal effects, the
conditional posterior should assign more mass to graphs and weights close to the
data-generating mechanism, and this concentration should increase as the
constraint set becomes more informative.
\subsubsection{Single-effect conditioning: validation on \texorpdfstring{$d=4$}{d=4}}
\label{sec:correctness_single_d4}
We construct 10 test cases by sampling 10 independent $d=4$ graphs and selecting
one ordered node pair from each graph. For each pair, we evaluate a one-sided
causal-effect tail query, represented in the score-level form
$\mathcal A_{\lambda_\star}=\{Z:h(Z)\ge\lambda_\star\}$ from
Sec.~\ref{sec:method}. For every test case, we run each stochastic method 10
times and report the mean estimate together with across-run variability.

\begin{figure}[t]
  \centering
  \includegraphics[width=1\linewidth,trim=1 1 1 1,clip]{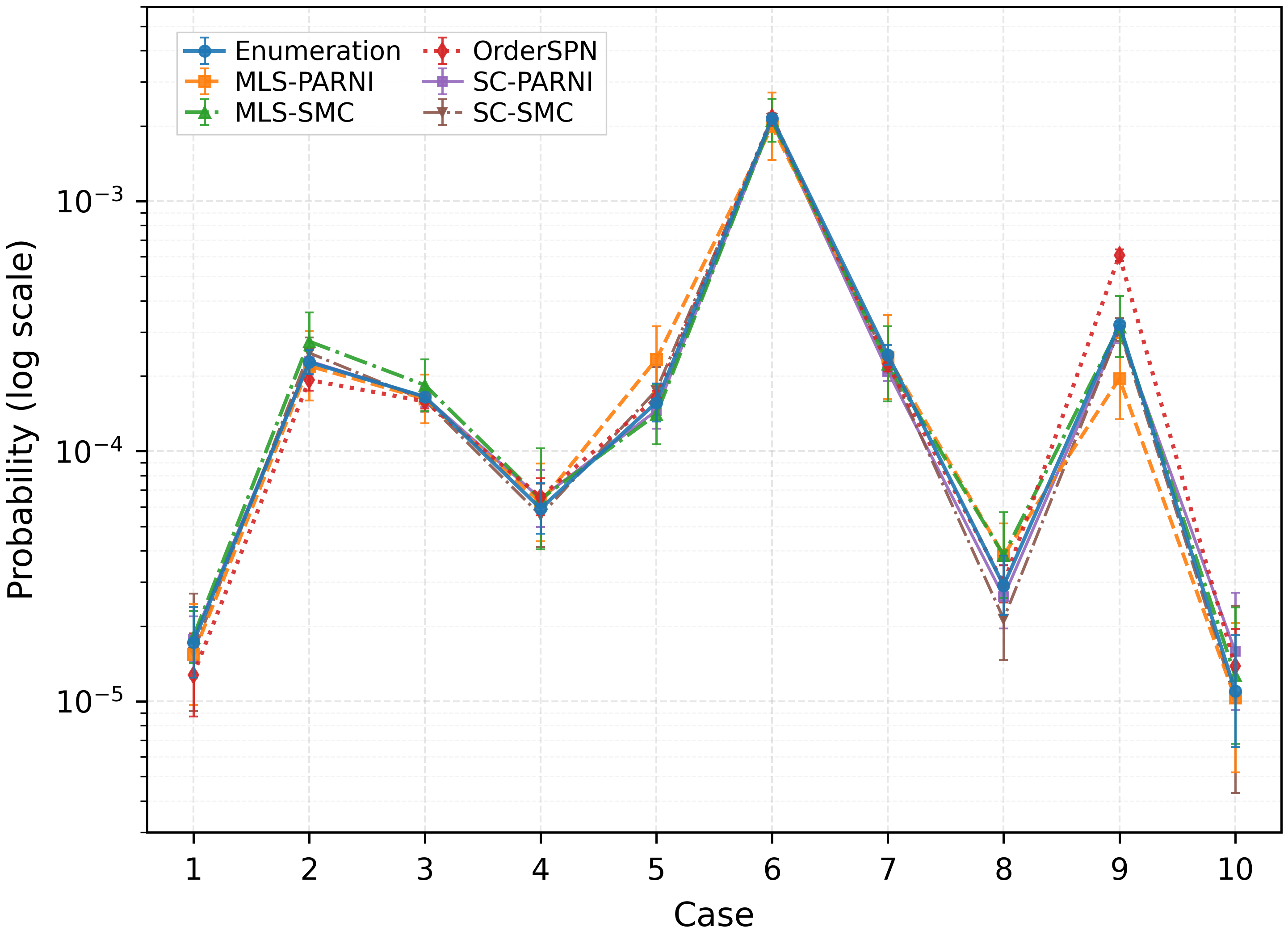}
  \caption{Single-effect one-sided tail-probability estimation on $d=4$ across
  10 test cases, with exhaustive enumeration as the reference posterior
  probability. Error bars show across-run variability over 10 independent runs.
  DiBS is omitted from the plot because its estimates are orders of magnitude
  larger on these tail events, which would obscure the comparison among the
  remaining methods; the corresponding numerical values are reported in
  Table~\ref{tab:d4_prob_10x7_mean_sd}.}
  \label{fig:single_ce_d4}
\end{figure}

Figure~\ref{fig:single_ce_d4} compares all methods against exhaustive
enumeration. We make two observations. First, all methods except DiBS are
broadly consistent with the enumeration baseline across the $d=4$ cases, with
only minor deviations; the corresponding numerical values are reported in
Table~\ref{tab:d4_prob_10x7_mean_sd}. Second, DiBS overestimates tail
probabilities by orders of magnitude on these tail events. We therefore exclude
DiBS from subsequent experiments and focus on methods that give reliable
small-graph estimates.

\subsubsection{Multi-effect conditioning: validating the conditioning operator on \texorpdfstring{$d=4$}{d=4} and \texorpdfstring{$d=8$}{d=8}}
\label{sec:correctness_multi}

The previous experiment validates tail-probability estimation for a single
causal-effect event. We next test whether the method correctly conditions the
joint posterior over graphs and weights. To make this test interpretable, we
construct constraints from the known data-generating state and ask whether the
resulting conditional posterior concentrates toward that state.

\paragraph{Multi-constraint conditioning event.}
We select $C$ ordered node pairs $\{(i_c,j_c)\}_{c=1}^{C}$. Let
$e_c^\star=\mathrm{CE}_{i_cj_c}(Z^\star)$ denote the ground-truth causal effect
for pair $(i_c,j_c)$ under the data-generating state
$Z^\star=(G^\star,B^\star)$. For each pair, we impose a two-sided interval
constraint around $e_c^\star$:
\begin{equation}
\label{eq:multi_ce_interval}
    \mathrm{CE}_{i_cj_c}(Z)
    \in
    [e_c^\star-\varepsilon,\,e_c^\star+\varepsilon],
    \quad c=1,\ldots,C.
\end{equation}
Each interval is represented by two one-sided inequalities, and the conjunction
of all inequalities defines the conditioning event. Following Sec.~\ref{sec:method},
we aggregate these inequalities using the minimum normalized margin score
$h_{\mathcal C}(Z)$, so that the target event is
$\mathcal E_{\mathcal C}=\{Z:h_{\mathcal C}(Z)\ge 0\}$.

We compare a \textbf{weak} constraint set, with fewer constrained pairs and hence
a looser event, against a \textbf{strong} constraint set, with more constrained
pairs and hence a tighter event. All posterior summaries are computed from the
final-stage multilevel-splitting particles and therefore condition on the target
event by construction. To visualize the conditional posterior, we aggregate
samples across runs to compute edge-frequency heatmaps, corresponding to
posterior marginal edge probabilities, and average edge-weight heatmaps,
corresponding to posterior mean weights. We also quantify structural recovery
using the Structural Hamming Distance (SHD) between sampled DAGs and the
ground-truth graph $G^\star$.

\begin{table}[t]
\centering
\caption{\textbf{$d=4$ (PARNI)} structural accuracy under multi-effect conditioning.}
\label{tab:d4_shd_parni}
\small
\begin{adjustbox}{width=\columnwidth, center}
\begin{tabular}{lrrrr}
\toprule
Constraints & Runs & Samples/run & Mean SHD $\downarrow$ & $\Pr(\mathrm{SHD}=0)\uparrow$ \\
\midrule
Weak   & 10 & 20 & $0.325 \pm 0.164$ & $0.790 \pm 0.117$ \\
Strong & 10 & 20 & $0.115 \pm 0.053$ & $0.885 \pm 0.053$ \\
\bottomrule
\end{tabular}
\end{adjustbox}
\end{table}

\begin{table}[t]
\centering
\caption{\textbf{$d=8$ (PARNI)} structural accuracy under multi-effect conditioning.
We report mean $\pm$ standard deviation of the per-run mean SHD, averaged over
sampled DAGs within each run, and the fraction of samples with $\mathrm{SHD}=0$.}
\label{tab:d8_shd_parni}
\small
\begin{adjustbox}{width=\columnwidth, center}
\begin{tabular}{lrrrr}
\toprule
Constraints & Runs & Samples/run & Mean SHD $\downarrow$ & $\Pr(\mathrm{SHD}=0)\uparrow$ \\
\midrule
Weak   & 10 & 50 & $5.342 \pm 0.442$ & $0.040 \pm 0.027$ \\
Strong & 10 & 50 & $2.290 \pm 0.199$ & $0.108 \pm 0.066$ \\
\bottomrule
\end{tabular}
\end{adjustbox}
\end{table}

\paragraph{Multi-effect conditioning results.}
As demonstrated by Figs.~\ref{fig:d4_heatmaps} and~\ref{fig:d8_heatmaps} in Appendix~\ref{sec:synthetic_heatmaps_app}, across both $d=4$ and $d=8$, stronger, more informative multi-effect constraints
consistently concentrate the conditional posterior toward the ground-truth
mechanism. Quantitatively, Tables~\ref{tab:d4_shd_parni} and~\ref{tab:d8_shd_parni}
show lower SHD under strong constraints than under weak constraints, together
with a higher fraction of exact structural matches. Qualitatively, the
corresponding edge-frequency and edge-weight heatmaps show the same pattern:
posterior mass becomes sharper and closer to the ground truth as the constraint
set becomes stronger. Together, these SHD improvements and posterior-summary
concentration patterns provide evidence that the framework correctly conditions
on multiple causal-effect constraints in the joint graph--parameter space.

\subsection{Scalability}
\label{sec:scalability}

The scalability experiments examine whether the rare-event estimator remains
stable when exact enumeration is no longer available. We focus on one-sided
single-effect tail-probability estimation on $d\in\{8,16,32\}$ using multilevel
splitting with the Structure-MCMC and PARNI-DAG kernels. These experiments test
three aspects of scalability: whether estimated tail probabilities decrease
smoothly as the target level $\lambda_\star$ becomes more stringent; whether
independent runs give reproducible tail curves; and whether informed structure
proposals become more important as the graph dimension increases.

\begin{figure}[t]
  \centering
  \includegraphics[
    width=0.95\linewidth,trim=5 5 5 5,clip
  ]{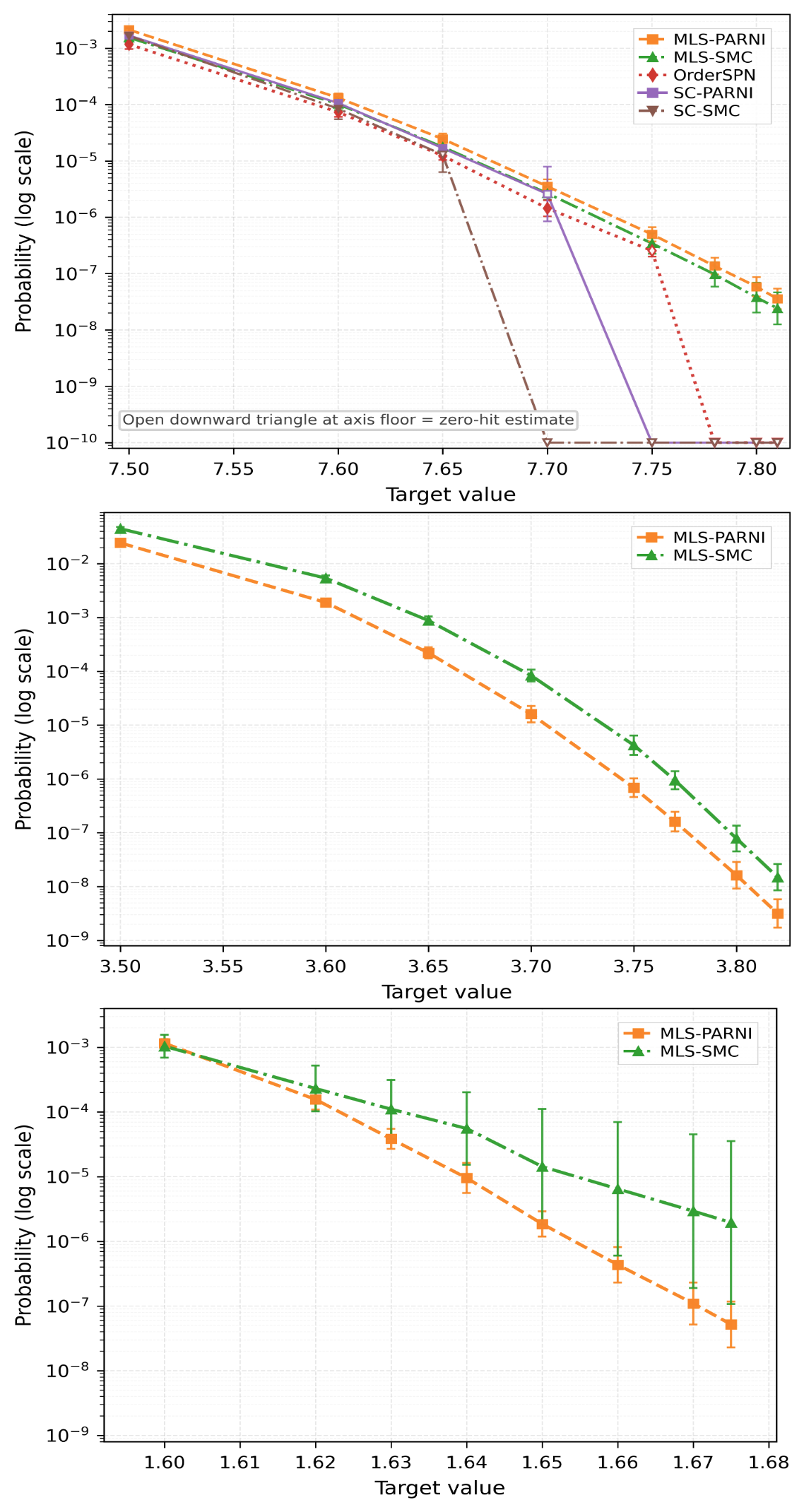}
  \caption{Single-effect one-sided tail-probability estimation on
  $d\in\{8,16,32\}$.}
  \label{fig:single_ce_d8_16_32}
\end{figure}

Figure~\ref{fig:single_ce_d8_16_32} summarizes one-sided single-effect tail-probability estimation on $d\in\{8,16,32\}$. On $d=8$, we additionally
include unconditional baselines, OrderSPN and two single-chain samplers, for
comparison. Both multilevel-splitting variants produce smooth, monotone-decaying
tail curves as the target level $\lambda_\star$ becomes more stringent. In
contrast, the unconditional methods quickly fail to generate samples satisfying
the more extreme targets, leading to degenerate probability estimates in the
rare-event region. This highlights the practical advantage of multilevel
splitting for conditional causal discovery on larger graphs, where rare-event
conditioning makes unconditional sampling increasingly inefficient.

The same curves also provide a threshold-sensitivity check. The final threshold
$\lambda_\star$ defines the scientific query, so the conditional posterior should
change as the threshold changes. Numerically, however, the estimated posterior
mass should vary smoothly and reproducibly as the threshold becomes more extreme.
Across independent runs, both structure kernels yield stable and monotone tail
curves in the moderate dimensions, supporting this expected behavior beyond the
enumerable $d=4$ setting.

Accordingly, for $d=16$ and $d=32$, we primarily assess robustness via internal
consistency, because exact gold-standard posterior enumeration is unavailable at
these scales. For $d=16$, both multilevel-splitting instantiations produce stable
and consistent probability estimates across runs, indicating that the framework
remains well behaved as graph size grows. At $d=32$, the two instantiations
separate more clearly: PARNI-DAG retains relatively stable run-to-run behavior,
whereas Structure-MCMC exhibits noticeably larger variability in the extreme
tail. This suggests that reliable deep-tail estimation in larger graphs benefits
from more informed structure proposals, as well as appropriate hyperparameter
settings.

\section{Case Study}\label{sec:sachs}
We now study an application of our conditional causal discovery framework to a real world protein-signalling dataset \citep{sachs2005causal} commonly used in causal discovery benchmarks. In particular, we will ask the following question: \emph{Which causal structures and directed pathways most plausibly realize a specified extreme-effect event?}

Rather than claiming new biological insights, our goal is to show that conditional sampling can identify and summarize plausible mechanisms in an extreme-effect region that is rare under the unconditional posterior.

\paragraph{Data, reference network, and target pairs.}
We use the Sachs protein-signaling dataset, containing \(n=7466\) measurements of \(d=11\) proteins \citep{sachs2005causal}. As a qualitative reference, we use the 20-edge consensus network in Fig.~\ref{fig:sachs_graph} \citep{koch2009learning,sachs2005causal}.
We study two target pairs:

(i) \(\mathrm{PIP3}\rightarrow \mathrm{PIP2}\) (indices \(6\to5\)), which admits a clean SEM decomposition into a direct route
\(\mathrm{PIP3}\to \mathrm{PIP2}\) and a mediated route \(\mathrm{PIP3}\to \mathrm{Plcg}\to \mathrm{PIP2}\);

(ii) \(\mathrm{Erk}\rightarrow \mathrm{Akt}\) (indices \(1\to10\)), a widely discussed cross-module influence in the Sachs signaling system.
\footnote{Node order used throughout:
[\texttt{Raf}, \texttt{Erk}, \texttt{Plcg}, \texttt{PKC}, \texttt{PKA}, \texttt{PIP2}, \texttt{PIP3},
\texttt{Mek}, \texttt{P38}, \texttt{Jnk}, \texttt{Akt}].
In particular, \(\mathrm{PIP3}\) is index 6,
\(\mathrm{PIP2}\) is index 5, \(\mathrm{Plcg}\) is index 2, \(\mathrm{Erk}\) is index 1, and \(\mathrm{Akt}\) is index 10.}

A published SEM mediation example on the same dataset reports a total effect of approximately \(0.5665\) for \(\mathrm{PIP3}\rightarrow\mathrm{PIP2}\), with an explicit direct/indirect decomposition via \(\mathrm{Plcg}\), which we use as an external numerical anchor for the effect scale \citep{madhanagopal2019analyzing}.
We treat such literature values as sanity checks on sign and magnitude rather than strict targets, since estimators and preprocessing differ.

\paragraph{Conditioning events.}
Using the notation from Sec.~\ref{sec:theory}, let $Z=(G,B)$ denote a graph--weight state sampled from the posterior \(\pi(Z)=p(G,B\mid \mathcal D)\), and let \(\mathrm{CE}_{ij}(Z)\) denote the signed linear-SEM total effect.
For the two target pairs, define the events $\mathcal E_{\mathrm{PIP}}(t)=\left\{Z:\mathrm{CE}_{\mathrm{PIP3},\mathrm{PIP2}}(Z)\ge t\right\}$,
and$\mathcal E_{\mathrm{ERK}}(t)=\left\{Z:\mathrm{CE}_{\mathrm{Erk},\mathrm{Akt}}(Z)\ge t\right\}$.
We compare four posterior conditions:
\begin{itemize}
    \item \textbf{Unconditioned:} \(Z\sim \pi(Z)=p(G,B\mid\mathcal D)\).
    \item \textbf{Cond-PIP:} \(Z\sim p(Z\mid\mathcal D,\mathcal E_{\mathrm{PIP}}(t_{\mathrm{PIP}}))\), with \(t_{\mathrm{PIP}}=0.77\).
    \item \textbf{Cond-ERK:} \(Z\sim p(Z\mid\mathcal D,\mathcal E_{\mathrm{ERK}}(t_{\mathrm{ERK}}))\), with \(t_{\mathrm{ERK}}=0.67\).
    \item \textbf{Cond-Joint:\\}
    \(Z\sim p\!\left(
            Z\mid\mathcal D,\,
            \mathcal E_{\mathrm{PIP}}(t'_{\mathrm{PIP}})
            \cap
            \mathcal E_{\mathrm{ERK}}(t'_{\mathrm{ERK}})
        \right),
    \)
    with \((t'_{\mathrm{PIP}},t'_{\mathrm{ERK}})=(0.74,0.65)\), targeting a similarly rare posterior region.
\end{itemize}
The estimated posterior masses of these events are reported in Appendix~\ref{sec:sachs_app} and Table~\ref{tab:sachs_event_probs}.
These conditioning events do \emph{not} assert that the dataset corresponds to a single ``ground-truth'' extreme state. Instead, they define posterior queries of the form $p(Z\mid \mathcal D,\mathcal E)$,
asking which graph--weight states and pathways remain plausible under the observed data when a specified causal effect, or a pair of causal effects, is unusually large.

\paragraph{Why unconditioned summaries can disagree with the consensus graph.}
We run experiments on each condition with settings listed in Sec.~\ref{sec:exp_setting}. Table~\ref{tab:sachs_4_effects} shows substantial unconditioned mass on zero total effect, corresponding to sampled DAGs with no directed path from the source to the target, even though the consensus network suggests nonzero coupling.
This is expected: the consensus graph is a qualitative reference rather than a uniquely identified ground truth, and the pooled perturbation data do not uniquely determine reachability under our model, so many near-equivalent posterior graphs omit these paths \citep{sachs2005causal,koch2009learning,friedman2003being}.

\paragraph{Mechanistic analysis under conditioning.}
\begin{itemize}
    \item \textbf{Cond-PIP:} The conditional posterior yields a highly concentrated explanation for \(\mathrm{PIP3}\to\mathrm{PIP2}\): the direct route and the \(\mathrm{Plcg}\)-mediated route appear in essentially all conditional samples, and the conditional mean effect is approximately \(0.776\) (calculated in Sec.~\ref{sec:sachs_app}), with an increased \(\mathrm{Plcg}\)-mediated share (Table~\ref{tab:sachs_canonical_paths}).

    \item \textbf{Cond-ERK:} \(\mathrm{Erk}\to\mathrm{Akt}\) remains direct-dominated, while \(\mathrm{PIP3}\to\mathrm{PIP2}\) becomes almost always reachable but is supported by many alternative, partly cancelling paths. This contrasts with the two-path concentration under Cond-PIP (Tables~\ref{tab:sachs_4_effects} and~\ref{tab:sachs_canonical_paths}).
    
    \item \textbf{Non-symmetry:} Conditioning on \(\mathrm{PIP3}\to\mathrm{PIP2}\) does not force \(\mathrm{Erk}\to\mathrm{Akt}\) to become extreme; for example, $\Pr\!\left(
            \mathrm{CE}_{\mathrm{Erk},\mathrm{Akt}}(Z)>0
            \mid \mathcal D,\mathcal E_{\mathrm{PIP}}(t_{\mathrm{PIP}})
        \right)
        \approx 0.498$.

    \item \textbf{Cond-Joint:} Joint conditioning increases the frequency of the mediated \(\mathrm{Erk}\to\mathrm{Plcg}\to\mathrm{Akt}\) route relative to Cond-ERK, highlighting \(\mathrm{Plcg}\) as a shared mediator under co-extreme coupling, beyond what either single-condition run reveals alone (Table~\ref{tab:sachs_canonical_paths}).
\end{itemize}

\paragraph{Graph-level shifts under conditioning.}
Figure~\ref{fig:sachs_4_heatmaps} illustrates these structural shifts, summarizing the edge frequencies and mean edge weights under the unconditioned posterior and the three conditional posteriors.

\paragraph{Takeaway.}
Without conditioning, posterior samples often imply zero or near-zero total effect, or distribute the effect across many paths that partially cancel each other, so pathway summaries are hard to determine.
Single-effect constraints make the dominant routes clearer: \textbf{Cond-PIP} explains \(\mathrm{PIP3}\to\mathrm{PIP2}\) mainly through two routes, whereas \textbf{Cond-ERK} keeps \(\mathrm{Erk}\to\mathrm{Akt}\) mostly direct but yields a broader set of \(\mathrm{PIP3}\to\mathrm{PIP2}\) paths with partial cancellation (Table~\ref{tab:sachs_canonical_paths}).
Joint conditioning is not just the overlap of the two single-effect results: compared to \textbf{Cond-ERK}, it more often highlights the mediated route \(\mathrm{Erk}\to\mathrm{Plcg}\to\mathrm{Akt}\), pointing to \(\mathrm{Plcg}\) as a shared mediator that emerges under the joint extreme-effect query (Table~\ref{tab:sachs_canonical_paths}).

\section{Conclusions}\label{sec:conclusion}
We introduced \emph{conditional causal discovery}, a framework for posterior
inference over causal graph structures and edge weights under user-specified
causal-effect constraints. The framework is designed to answer two questions
simultaneously: how likely a specified extreme-effect event is under the
posterior, and which graph--weight configurations remain plausible when that
event occurs.

To make such inference practical in rare-event regions, we developed an adaptive
multilevel splitting framework with an MCMC kernel over the joint graph--weight
space. By combining structure proposals, such as PARNI-DAG or Structure-MCMC,
with blocked weight moves, the method estimates one-sided signed causal-effect
tail probabilities and produces representative samples from the corresponding
constrained posterior.

Empirically, we validated the estimator against exhaustive enumeration in a
four-node setting and showed that the approach remains effective as problem size
grows, where unconditional baselines often degenerate in the rare-event region.
In a case study with the Sachs dataset, conditional samples provided compact
pathway-level explanations and highlighted coherent mechanisms under both
single-effect and joint-effect queries.

\paragraph{Limitations and future work.}
Our current implementation assumes linear-Gaussian SEMs under causal sufficiency.
Extending the framework to nonlinear mechanisms, latent confounding, and
interventional data remains an important direction. At the algorithmic level, the
multilevel splitting procedure only requires a scalar score function defining the
event of interest. For nonlinear structural causal models, the main changes would
be to replace the BGe and conjugate posterior components with suitable nonlinear
posterior inference modules, and to replace the closed-form linear causal-effect
evaluator with an appropriate effect-estimation procedure. More broadly,
incorporating richer constraint families, such as path-specific effects or
qualitative monotonicity constraints, could enable deeper mechanistic analysis of
complex scientific systems.

\FloatBarrier 
\clearpage
\bibliography{uai2026-template}
\newpage
\clearpage
\newpage

\appendix
\section{Appendix}

\subsection{Experiment Setting}
\label{sec:exp_setting}
For all synthetic experiments, we generate the ground-truth DAG $G^\star$ using
an ordered Erd\H{o}s--R\'enyi construction. Given dimension $d$ and edge budget
parameter $\texttt{edges\_per\_node}=2$, we set
\[
    p_{\mathrm{edge}}
    =
    \min\!\left\{
    \frac{\texttt{edges\_per\_node}\cdot d}{d(d-1)/2},\,0.5
    \right\}.
\]
We then sample a Bernoulli adjacency matrix with edge probability
$p_{\mathrm{edge}}$, keep only the strict lower-triangular part to ensure
acyclicity under the current ordering, and finally apply a random permutation of
the node labels. The effective edge probabilities are therefore
$p_{\mathrm{edge}}=0.5$ for $d=4$, $p_{\mathrm{edge}}=0.5$ for $d=8$,
$p_{\mathrm{edge}}=4/15$ for $d=16$, and $p_{\mathrm{edge}}=4/31$ for
$d=32$. The cap at $0.5$ is the small-graph handling used by the generator: for
$d=4$, the nominal value $4/3$ is clipped to $0.5$, giving an expected $3$
edges instead of an overly dense four-node DAG. The corresponding expected edge
counts are $3$, $14$, $32$, and $64$ for $d=4,8,16,32$, respectively.

Conditional on $G^\star$, absent edges are assigned weight zero and nonzero edge
weights are sampled independently from a Gaussian distribution, as in the data
generation script. Observations are then generated from the linear-Gaussian SEM
using $n_{\mathrm{obs}}=1000$ training samples.

Below are the experiment settings for each method:
\begin{enumerate}
 \item \textit{Exhaustive enumeration ($d=4$ only):}
 enumerate all DAGs ($|\mathcal G_{d=4}| = 543$) and draw $20000$ weight samples per graph.
 \item \textit{OrderSPN \citep{wang2022tractable}:}
 sample $5000$ graphs from the learned circuit and draw $100$ weight samples per graph.
 \item \textit{DiBS \citep{lorch2021dibs}:}
 draw $500000$ joint graph--weight samples in total.
 \item \textit{PARNI-DAG single chain \citep{liang2023structure}:}
 run a single long MCMC chain over $(G,B)$ using PARNI-DAG as the structure proposal kernel for $500000$ iterations with $10\%$ burn-in.
 \item \textit{Structure-MCMC single chain:}
 run a single long MCMC chain over $(G,B)$ using Structure-MCMC as the structure proposal kernel for $500000$ iterations with $10\%$ burn-in.
 \item \textit{MLS-PARNI-DAG \citep{liang2023structure}:}
 use our MLS framework with PARNI-DAG as the structure proposal kernel. We use $N=200$ particles, $m=2000$ MCMC iterations per level, and $K_{\max}=10$ levels for all dimensions.
 \item \textit{MLS-Structure-MCMC:}
 use our MLS framework with Structure-MCMC as the structure proposal kernel, with the same MLS hyperparameters as above.
\end{enumerate}

\subsection{Target-threshold construction}
\label{sec:target_threshold_construction}

We construct target score levels to evaluate each method's ability to estimate
\emph{deep-tail} probabilities under increasingly extreme one-sided
causal-effect constraints. Let
\[
    e_0
    =
    \mathrm{CE}_{ij}(Z^\star)
    =
    \mathrm{CE}_{ij}(G^\star,B^\star)
\]
denote the ground-truth signed total causal effect for a queried ordered pair
$(i,j)$, where $Z^\star=(G^\star,B^\star)$ is the data-generating graph--weight
state.

For synthetic experiments, we choose the tail direction using the sign of
$e_0$. Equivalently, define $s=\operatorname{sign}(e_0)$ and use the scalar
score
\[
    h(Z)=s\,\mathrm{CE}_{ij}(Z).
\]
Then larger values of $h(Z)$ always correspond to more extreme effects in the
selected direction. A target score level $\lambda$ defines the event
\[
    \mathcal A_\lambda
    =
    \{Z:h(Z)\ge \lambda\}.
\]
When $e_0>0$, this is the right-tail event
$\mathcal E_{ij}^{+}(\lambda)=\{Z:\mathrm{CE}_{ij}(Z)\ge \lambda\}$. When
$e_0<0$, it is the left-tail event
$\mathcal E_{ij}^{-}(\lambda)=\{Z:\mathrm{CE}_{ij}(Z)\le -\lambda\}$. This is
the same score-level representation used by the MLS sampler in
Sec.~\ref{sec:method}.

\begin{enumerate}
  \item \textbf{Initial score level.}
        We initialize the threshold grid at the magnitude of the ground-truth
        effect, $\lambda_1=|e_0|$.

  \item \textbf{Pilot run.}
        We run a short, low-budget adaptive multilevel-splitting pilot to
        identify a more extreme target score level $\lambda_T$ such that the
        corresponding tail probability is already in the rare-event regime,
        approximately $10^{-6}$ or smaller.

  \item \textbf{Threshold grid.}
        We form a monotone increasing sequence of $T$ target score levels
        $\lambda_1<\lambda_2<\cdots<\lambda_T$. Larger $\lambda$ always
        corresponds to a more extreme one-sided effect in the selected
        direction.
\end{enumerate}

\subsection{Detailed result for d=4 baseline experiment}
\newcolumntype{Y}{>{\centering\arraybackslash}X}
Table~\ref{tab:d4_prob_10x7_mean_sd} lists statistics for the $d=4$ baseline experiments across seven methods.
\begin{table*}[tbp]
\centering
\caption{$d=4$ per-case probability estimates across runs (mean $\pm$ SD). MLS refers to multilevel splitting, and SC refers to single-chain sampling.}
\label{tab:d4_prob_10x7_mean_sd}
\scriptsize
\setlength{\tabcolsep}{2.5pt}
\renewcommand{\arraystretch}{1.15}

\begin{tabularx}{\linewidth}{c*{7}{Y}}
\toprule
Case &
\makecell{\textbf{Enumeration}\\$\hat p\pm\mathrm{SD}$} &
\makecell{\textbf{PARNI-MLS}\\$\hat p\pm\mathrm{SD}$} &
\makecell{\textbf{Structure-MCMC-MLS}\\$\hat p\pm\mathrm{SD}$} &
\makecell{\textbf{SPN}\\$\hat p\pm\mathrm{SD}$} &
\makecell{\textbf{PARNI-SC}\\$\hat p\pm\mathrm{SD}$} &
\makecell{\textbf{Structure-MCMC-SC}\\$\hat p\pm\mathrm{SD}$} &
\makecell{\textbf{DIBS}\\$\hat p\pm\mathrm{SD}$} \\
\midrule
1 & \makecell{$ 1.87\times 10^{-5} $\\{\tiny$\pm\,7.25\times 10^{-6}$}} & \makecell{$ 1.83\times 10^{-5} $\\{\tiny$\pm\,1.07\times 10^{-5}$}} & \makecell{$ 1.9\times 10^{-5} $\\{\tiny$\pm\,6.09\times 10^{-6}$}} & \makecell{$ 1.42\times 10^{-5} $\\{\tiny$\pm\,6\times 10^{-6}$}} & \makecell{$ 1.84\times 10^{-5} $\\{\tiny$\pm\,5.32\times 10^{-6}$}} & \makecell{$ 1.4\times 10^{-5} $\\{\tiny$\pm\,1.01\times 10^{-5}$}} & \makecell{$ 2.32\times 10^{-2} $\\{\tiny$\pm\,1.35\times 10^{-3}$}} \\
2 & \makecell{$ 2.3\times 10^{-4} $\\{\tiny$\pm\,1.13\times 10^{-4}$}} & \makecell{$ 2.5\times 10^{-4} $\\{\tiny$\pm\,1.07\times 10^{-4}$}} & \makecell{$ 2.41\times 10^{-4} $\\{\tiny$\pm\,8.04\times 10^{-5}$}} & \makecell{$ 1.37\times 10^{-2} $\\{\tiny$\pm\,8.62\times 10^{-4}$}} & \makecell{$ 2.28\times 10^{-4} $\\{\tiny$\pm\,7.49\times 10^{-5}$}} & \makecell{$ 2.53\times 10^{-4} $\\{\tiny$\pm\,5.78\times 10^{-5}$}} & \makecell{$ 1.35\times 10^{-2} $\\{\tiny$\pm\,7.72\times 10^{-4}$}} \\
3 & \makecell{$ 1.65\times 10^{-4} $\\{\tiny$\pm\,5.96\times 10^{-5}$}} & \makecell{$ 1.7\times 10^{-4} $\\{\tiny$\pm\,5.47\times 10^{-5}$}} & \makecell{$ 1.61\times 10^{-4} $\\{\tiny$\pm\,1.41\times 10^{-5}$}} & \makecell{$ 1.83\times 10^{-1} $\\{\tiny$\pm\,7.92\times 10^{-3}$}} & \makecell{$ 1.67\times 10^{-4} $\\{\tiny$\pm\,2.42\times 10^{-5}$}} & \makecell{$ 1.79\times 10^{-4} $\\{\tiny$\pm\,2.31\times 10^{-5}$}} & \makecell{$ 1.81\times 10^{-1} $\\{\tiny$\pm\,8.38\times 10^{-3}$}} \\
4 & \makecell{$ 6.17\times 10^{-5} $\\{\tiny$\pm\,4.07\times 10^{-5}$}} & \makecell{$ 5.89\times 10^{-5} $\\{\tiny$\pm\,3.67\times 10^{-5}$}} & \makecell{$ 6.79\times 10^{-5} $\\{\tiny$\pm\,2.13\times 10^{-5}$}} & \makecell{$ 2.67\times 10^{-2} $\\{\tiny$\pm\,1.76\times 10^{-3}$}} & \makecell{$ 7.06\times 10^{-5} $\\{\tiny$\pm\,2.85\times 10^{-5}$}} & \makecell{$ 7.21\times 10^{-5} $\\{\tiny$\pm\,3.06\times 10^{-5}$}} & \makecell{$ 2.65\times 10^{-2} $\\{\tiny$\pm\,1.66\times 10^{-3}$}} \\
5 & \makecell{$ 1.6\times 10^{-4} $\\{\tiny$\pm\,5.56\times 10^{-5}$}} & \makecell{$ 1.71\times 10^{-4} $\\{\tiny$\pm\,5.9\times 10^{-5}$}} & \makecell{$ 1.62\times 10^{-4} $\\{\tiny$\pm\,1.74\times 10^{-5}$}} & \makecell{$ 3.34\times 10^{-2} $\\{\tiny$\pm\,1.88\times 10^{-3}$}} & \makecell{$ 1.55\times 10^{-4} $\\{\tiny$\pm\,4.54\times 10^{-5}$}} & \makecell{$ 1.69\times 10^{-4} $\\{\tiny$\pm\,4.88\times 10^{-5}$}} & \makecell{$ 3.34\times 10^{-2} $\\{\tiny$\pm\,1.63\times 10^{-3}$}} \\
6 & \makecell{$ 2.15\times 10^{-3} $\\{\tiny$\pm\,7.08\times 10^{-4}$}} & \makecell{$ 2.21\times 10^{-3} $\\{\tiny$\pm\,9.14\times 10^{-4}$}} & \makecell{$ 2.16\times 10^{-3} $\\{\tiny$\pm\,7.4\times 10^{-4}$}} & \makecell{$ 2.27\times 10^{-2} $\\{\tiny$\pm\,1.61\times 10^{-3}$}} & \makecell{$ 2.09\times 10^{-3} $\\{\tiny$\pm\,1.38\times 10^{-4}$}} & \makecell{$ 2.15\times 10^{-3} $\\{\tiny$\pm\,1.78\times 10^{-4}$}} & \makecell{$ 2.3\times 10^{-2} $\\{\tiny$\pm\,1.55\times 10^{-3}$}} \\
7 & \makecell{$ 2.44\times 10^{-4} $\\{\tiny$\pm\,1.13\times 10^{-4}$}} & \makecell{$ 2.4\times 10^{-4} $\\{\tiny$\pm\,1.06\times 10^{-4}$}} & \makecell{$ 2.19\times 10^{-4} $\\{\tiny$\pm\,2.48\times 10^{-5}$}} & \makecell{$ 9.71\times 10^{-2} $\\{\tiny$\pm\,6.54\times 10^{-3}$}} & \makecell{$ 2.22\times 10^{-4} $\\{\tiny$\pm\,2.25\times 10^{-5}$}} & \makecell{$ 2.42\times 10^{-4} $\\{\tiny$\pm\,2.5\times 10^{-5}$}} & \makecell{$ 9.65\times 10^{-2} $\\{\tiny$\pm\,6.06\times 10^{-3}$}} \\
8 & \makecell{$ 3.09\times 10^{-5} $\\{\tiny$\pm\,2.84\times 10^{-5}$}} & \makecell{$ 4.48\times 10^{-5} $\\{\tiny$\pm\,2.71\times 10^{-5}$}} & \makecell{$ 3.1\times 10^{-5} $\\{\tiny$\pm\,6.72\times 10^{-6}$}} & \makecell{$ 5.76\times 10^{-3} $\\{\tiny$\pm\,6.65\times 10^{-4}$}} & \makecell{$ 2.82\times 10^{-5} $\\{\tiny$\pm\,8.1\times 10^{-6}$}} & \makecell{$ 3.4\times 10^{-5} $\\{\tiny$\pm\,1.35\times 10^{-5}$}} & \makecell{$ 5.86\times 10^{-3} $\\{\tiny$\pm\,5.4\times 10^{-4}$}} \\
9 & \makecell{$ 3.22\times 10^{-4} $\\{\tiny$\pm\,1.34\times 10^{-4}$}} & \makecell{$ 3.25\times 10^{-4} $\\{\tiny$\pm\,1.33\times 10^{-4}$}} & \makecell{$ 3.22\times 10^{-4} $\\{\tiny$\pm\,5.19\times 10^{-5}$}} & \makecell{$ 1.32\times 10^{-2} $\\{\tiny$\pm\,1.46\times 10^{-3}$}} & \makecell{$ 3.09\times 10^{-4} $\\{\tiny$\pm\,6.17\times 10^{-5}$}} & \makecell{$ 3.3\times 10^{-4} $\\{\tiny$\pm\,4.04\times 10^{-5}$}} & \makecell{$ 1.33\times 10^{-2} $\\{\tiny$\pm\,1.45\times 10^{-3}$}} \\
10 & \makecell{$ 1.33\times 10^{-5} $\\{\tiny$\pm\,1.02\times 10^{-5}$}} & \makecell{$ 1.98\times 10^{-5} $\\{\tiny$\pm\,1.78\times 10^{-5}$}} & \makecell{$ 1.37\times 10^{-5} $\\{\tiny$\pm\,1.11\times 10^{-5}$}} & \makecell{$ 1.82\times 10^{-2} $\\{\tiny$\pm\,1.55\times 10^{-3}$}} & \makecell{$ 1.68\times 10^{-5} $\\{\tiny$\pm\,1.52\times 10^{-5}$}} & \makecell{$ 1.21\times 10^{-5} $\\{\tiny$\pm\,1.58\times 10^{-5}$}} & \makecell{$ 1.8\times 10^{-2} $\\{\tiny$\pm\,1.72\times 10^{-3}$}} \\
\bottomrule
\end{tabularx}
\end{table*}

\subsection{Time Complexity}
\label{sec:time_complex}

We separate the cost of the outer adaptive multilevel-splitting loop from the
cost of the inner MCMC mutation kernel. Let $N$ be the number of particles,
$m$ the number of MCMC mutation steps per particle per level, $K$ the number of
splitting levels, and $C_{\mathrm{MH}}$ the average cost of one Metropolis--
Hastings proposal, including proposal generation, posterior-ratio evaluation,
and causal-effect score evaluation. At each level, the algorithm mutates
$N$ particles for $m$ steps, computes $N$ scores, sorts or partially sorts the
scores to choose the next empirical quantile, and resamples the survivors. The
overall cost is therefore
\[
    O\!\left(KNmC_{\mathrm{MH}} + KN\log N\right),
\]
where the $KN\log N$ term comes from quantile selection. If selection is
implemented by a linear-time order-statistic routine, this sorting term can be
reduced to $O(KN)$ and the mutation cost dominates.

The average proposal cost depends on the mixture of structure and weight moves.
If $p_{\mathrm{struct}}$ is the probability of proposing a structure move, then
\[
    C_{\mathrm{MH}}
    \approx
    p_{\mathrm{struct}} C_{\mathrm{struct}}
    +
    (1-p_{\mathrm{struct}}) C_{\mathrm{weight}}
    +
    C_h,
\]
where $C_{\mathrm{struct}}$ is the cost of proposing and scoring a graph update,
$C_{\mathrm{weight}}$ is the cost of a blocked coefficient refresh, and $C_h$ is
the cost of evaluating the scalar score $h(Z)$. In the linear-Gaussian
implementation, the coefficient refreshes use node-wise conjugate posterior
updates, while $h(Z)$ is computed from the linear total-effect matrix. A direct
matrix inverse gives $C_h=O(d^3)$, although for a DAG and a single target pair
one can exploit the topological structure or solve a triangular system to reduce
this cost in practice.

The key rare-event advantage is the dependence on the event probability
$p_\star$. A direct posterior sampler needs about $1/p_\star$ samples to see one
sample from an event of posterior probability $p_\star$, and $O(1/(r^2p_\star))$
samples to estimate that probability with fixed relative error $r$. By contrast,
adaptive multilevel splitting replaces the single rare event with a product of
moderate conditional survival probabilities. If the empirical survival fraction
is approximately $\rho$ at each level, then $K$ grows roughly like
$\log(p_\star)/\log(\rho)$, so the leading cost grows approximately
logarithmically in $1/p_\star$ rather than linearly in $1/p_\star$. This is why
multilevel splitting remains useful in the deep-tail regimes where unconditional
sampling degenerates.

\paragraph{Runtime measurements.}
Table~\ref{tab:runtime_d4} reports wall-clock running times for the $d=4$ case 1
experiment. Table~\ref{tab:outer_mls_runtime} reports the average time per outer
MLS loop for the two structure kernels across dimensions. PARNI-DAG is more
expensive per loop because its locally informed proposal requires additional
neighborhood construction and scoring, while Structure-MCMC has cheaper local
edge proposals. The benefit of PARNI-DAG is not lower per-iteration cost, but
more stable deep-tail behavior in larger graphs, as shown in the scalability
experiments.

\begin{table}[t]
\centering
\caption{Average running time for the $d=4$ case 1 experiment over 10 runs. We report mean $\pm$ standard deviation in seconds.}
\label{tab:runtime_d4}
\begin{tabular}{lc}
\toprule
Method & Time (s) \\
\midrule
MLS-PARNI-DAG      & $192.3 \pm 4.8$ \\
MLS-Structure-MCMC & $56.1 \pm 2.1$ \\
PARNI-DAG          & $268.2 \pm 5.3$ \\
Structure-MCMC     & $105.5 \pm 3.0$ \\
Enumeration-Baseline & $301.6 \pm 4.6$ \\
OrderSPN           & $166.8 \pm 3.9$ \\
DiBS               & $434.2 \pm 6.7$ \\
\bottomrule
\end{tabular}
\end{table}

\begin{table}[t]
\centering
\caption{Average running time per outer MLS loop in seconds. Standard deviations over 10 runs are shown in parentheses.}
\label{tab:outer_mls_runtime}
\resizebox{\columnwidth}{!}{%
\begin{tabular}{lcccc}
\toprule
Kernel & $d=4$ & $d=8$ & $d=16$ & $d=32$ \\
\midrule
PARNI-DAG      & 38.4 (1.1) & 56.8 (1.6) & 92.4 (2.8) & 172.5 (4.9) \\
Structure-MCMC & 11.2 (0.5) & 19.5 (0.8) & 34.8 (1.2) & 58.5 (2.1) \\
\bottomrule
\end{tabular}%
}
\end{table}

\subsection{Extension to nonlinear Gaussian mechanisms}
\label{sec:nonlinear_extension}

The outer MLS framework is not specific to linear-Gaussian SEMs. It only
requires three ingredients: an unconstrained posterior target $\pi(Z)$, a scalar
score $h(Z)$ whose super-level set defines the event of interest, and an MCMC
kernel that approximately preserves the level-truncated posterior
$\pi(Z)\mathbf 1\{h(Z)\ge\lambda\}$. The linear-Gaussian assumptions used in the
main experiments provide convenient closed forms for the BGe marginal likelihood,
the node-wise coefficient posterior, and the total causal effect, but they are
not required by the splitting principle itself.

For a nonlinear Gaussian structural causal model, one could write
\[
    X_j = f_j\!\left(X_{\mathrm{pa}_G(j)};\phi_j\right)+\epsilon_j,
    \qquad
    \epsilon_j\sim\mathcal N(0,\sigma_j^2),
\]
and replace the graph--weight state $Z=(G,B)$ by a graph--mechanism state
$Z=(G,\phi,\sigma)$. The posterior target would become
$p(G,\phi,\sigma\mid\mathcal D)$, obtained using an appropriate nonlinear
mechanism class such as splines, Gaussian processes, or neural networks. The
inner mutation kernel would then update graph structure and mechanism parameters
instead of graph structure and linear coefficients.

The causal-effect score would also be replaced by a nonlinear effect evaluator.
For example, for a scalar intervention one may define an average interventional
contrast,
\[
    h(Z)
    =
    \mathbb E_Z\!\left[X_j\mid \mathrm{do}(X_i=x+\Delta)\right]
    -
    \mathbb E_Z\!\left[X_j\mid \mathrm{do}(X_i=x)\right],
\]
or an averaged derivative when such derivatives are well-defined. These
expectations can be estimated by ancestral simulation under the proposed
nonlinear SCM. The outer splitting loop would remain unchanged; the additional
computational burden would come from evaluating nonlinear posterior ratios and
interventional scores. Therefore, the main technical requirement for such an
extension is not a new rare-event algorithm, but reliable posterior inference and
causal-effect evaluation for the chosen nonlinear mechanism class.

\subsection{Sachs experiment calculations and tables}
\label{sec:sachs_app}

This section provides supplementary numerical details for the Sachs case study in Sec.~\ref{sec:sachs}, including
(i) the qualitative reference network,
(ii) estimated posterior masses of the tail events, and
(iii) pathway-level decompositions of conditional effects.

\begin{figure}[t]
  \centering
  \includegraphics[width=0.7\linewidth,trim=8 10 8 10,clip]{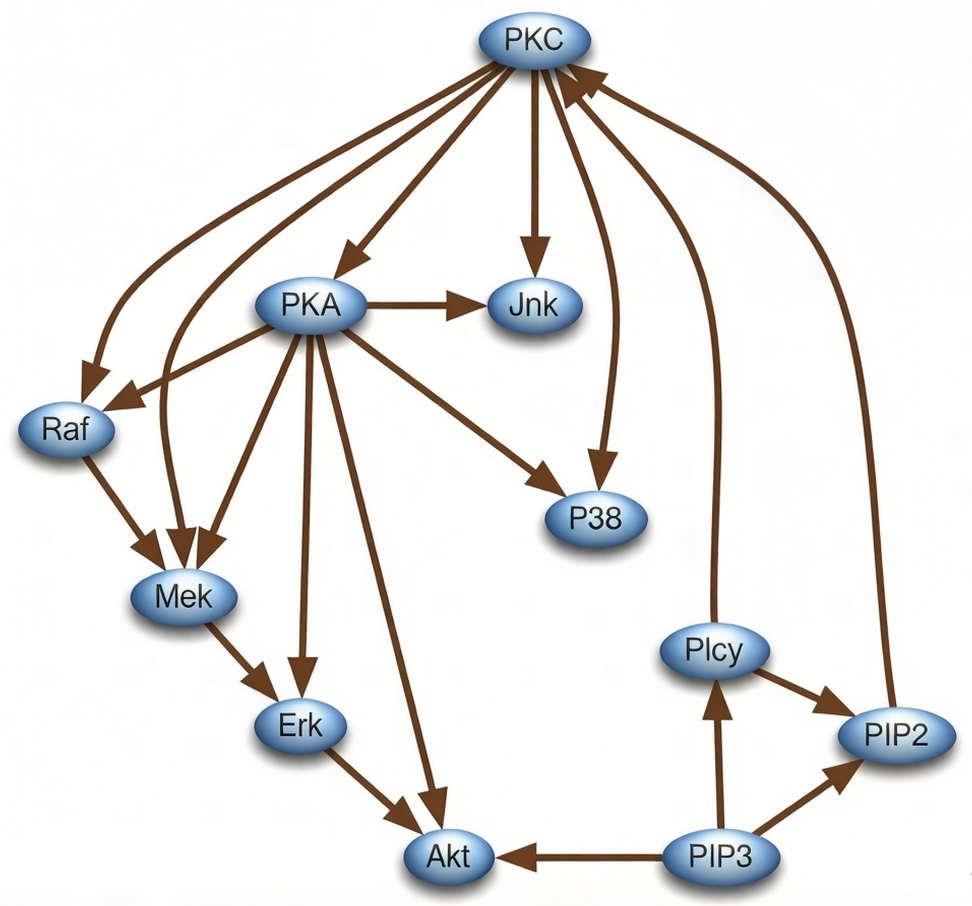}
  \caption{Consensus signaling network (Sachs) used as a qualitative reference.}
  \label{fig:sachs_graph}
\end{figure}

\paragraph{Reference network.}
Figure~\ref{fig:sachs_graph} shows the 11-node, 20-edge consensus signaling network used as a qualitative reference in Sec.~\ref{sec:sachs}.

\paragraph{Posterior mass of tail events.}
We run MLS with 10 independent repetitions for each conditioning regime.
Table~\ref{tab:sachs_event_probs} reports (i) the \emph{geometric mean} estimate of $\Pr(\mathcal{E}\mid\mathcal{D})$
(i.e., $\exp(\mathbb{E}[\log \hat p])$ across runs) and
(ii) the mean $\pm$ sd of $-\log \hat p$ across runs.

\begin{table}[!htbp]
\centering
\small
\caption{Estimated posterior mass of Sachs tail events over 10 MLS runs.}
\label{tab:sachs_event_probs}
\setlength{\tabcolsep}{4pt}
\begin{tabular}{lcc}
\toprule
Condition & $\Pr(\mathcal{E}\mid\mathcal{D})$ (geo-mean) & $-\log \hat{p}$ (mean $\pm$ sd) \\
\midrule
Cond-PIP   & $2.2 \times 10^{-7}$ & $15.32 \pm 0.44$ \\
Cond-ERK   & $1.3 \times 10^{-7}$ & $15.84 \pm 0.53$ \\
Cond-Joint & $5.2 \times 10^{-7}$ & $14.47 \pm 0.67$ \\
\bottomrule
\end{tabular}
\end{table}

\paragraph{Path decomposition under Cond-PIP.}
Across samples drawn from the conditional posterior
\[
    p\!\left(Z\mid \mathcal D,\mathcal E_{\mathrm{PIP}}(t_{\mathrm{PIP}})\right),
\]
the two canonical pathways
$\mathrm{PIP3}\!\to\!\mathrm{PIP2}$ and
$\mathrm{PIP3}\!\to\!\mathrm{Plcg}\!\to\!\mathrm{PIP2}$
are present in \emph{100\%} of samples.
These two routes account for essentially the entire conditional mean effect:
\begin{equation}
\begin{aligned}
&\mathbb{E}\!\left[
\mathrm{CE}_{\mathrm{PIP3},\mathrm{PIP2}}(Z)
\mid \mathcal{D}, \mathcal{E}_{\mathrm{PIP}}(t_{\mathrm{PIP}})
\right] \\
&\qquad \approx
\underbrace{0.6142}_{\text{direct}}
+
\underbrace{0.1614}_{\text{via Plcg}}
+
\underbrace{2\cdot 10^{-4}}_{\text{other paths}}
\approx 0.7758.
\end{aligned}
\end{equation}

The contribution from all remaining directed paths is numerically negligible,
confirming that conditioning isolates a highly concentrated two-path mechanism.

\paragraph{Summary tables.}
Table~\ref{tab:sachs_4_effects} reports posterior effect statistics (e.g., $\Pr(\mathrm{CE}>0)$, conditional means, and path counts) across the four regimes.
Table~\ref{tab:sachs_canonical_paths} further decomposes the conditional mean effects into direct and mediated contributions for the two target pairs (PIP3$\to$PIP2 and Erk$\to$Akt), together with the frequency of the canonical mediated paths.

\begin{figure*}[tbp]
  \centering
  \captionsetup{font=small}
  \captionsetup[subfigure]{labelformat=simple}
  \renewcommand\thesubfigure{(\alph{subfigure})}
  \captionsetup[sub]{font=footnotesize}
  \begin{subfigure}[t]{0.49\textwidth}
    \centering
    \includegraphics[width=\linewidth]{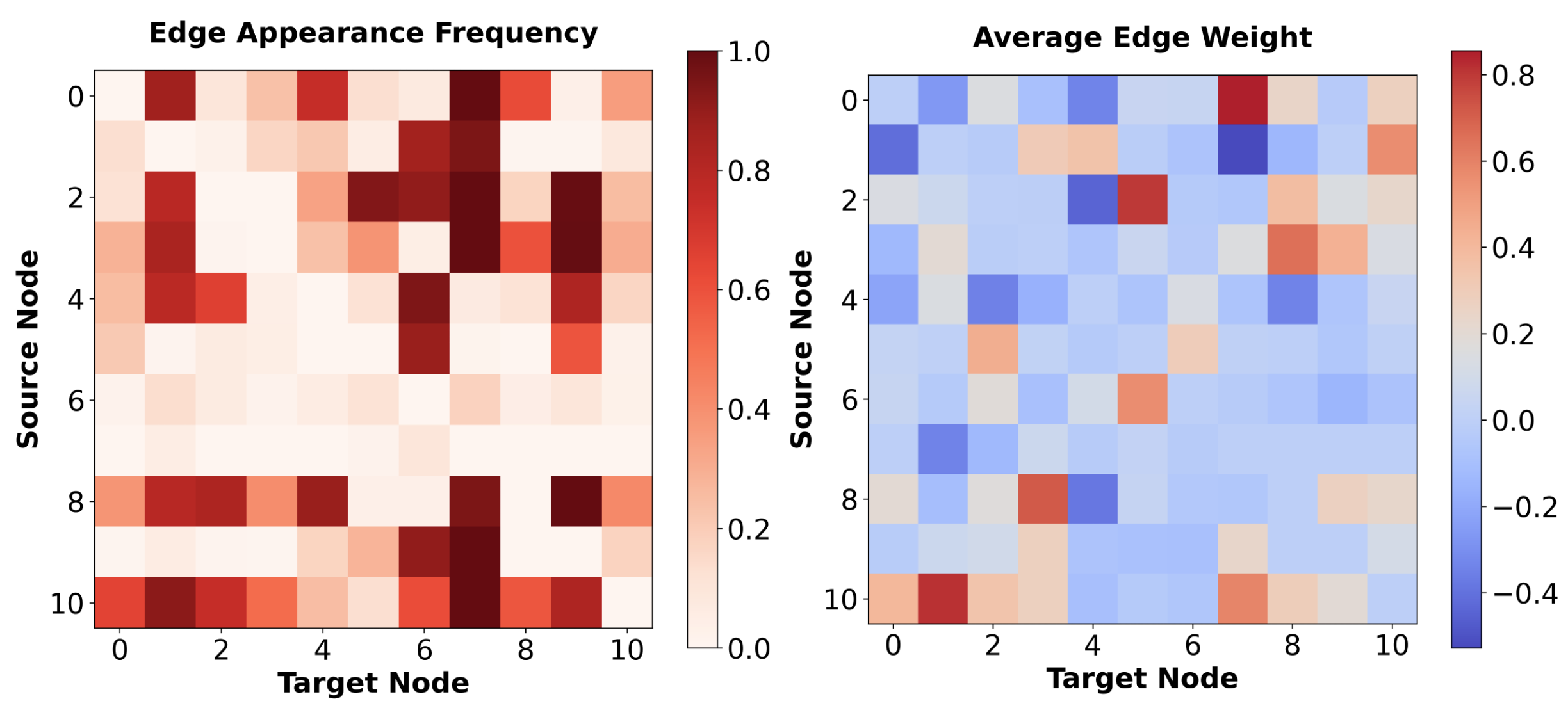}
    \caption{Unconditioned posterior}
    \label{fig:sachs_hm_uncond}
  \end{subfigure}\hfill
  \begin{subfigure}[t]{0.49\textwidth}
    \centering
    \includegraphics[width=\linewidth]{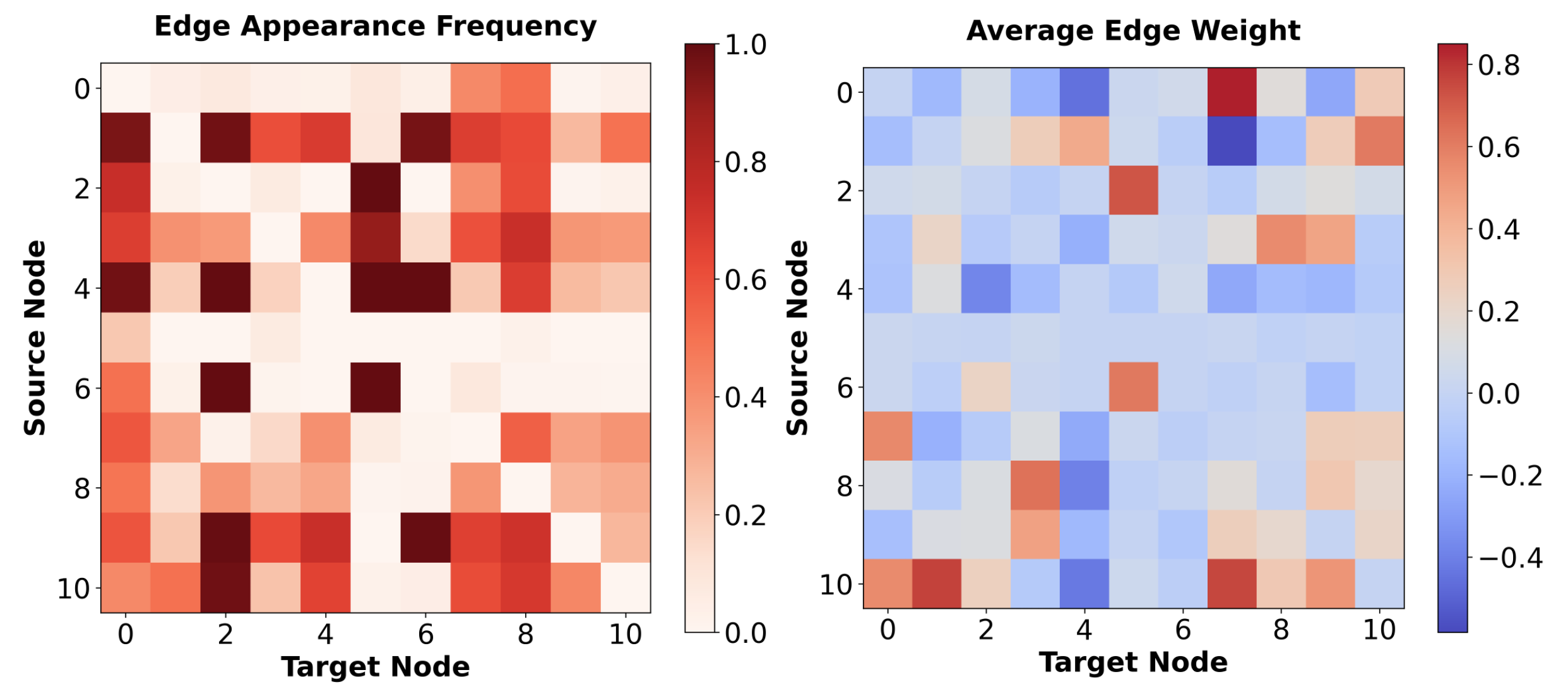}
    \caption{Cond-PIP: $\mathrm{CE}_{\mathrm{PIP3}\to\mathrm{PIP2}}>t_{\mathrm{PIP}}$}
    \label{fig:sachs_hm_cond_pip}
  \end{subfigure}
  \vspace{2mm}
  \begin{subfigure}[t]{0.49\textwidth}
    \centering
    \includegraphics[width=\linewidth]{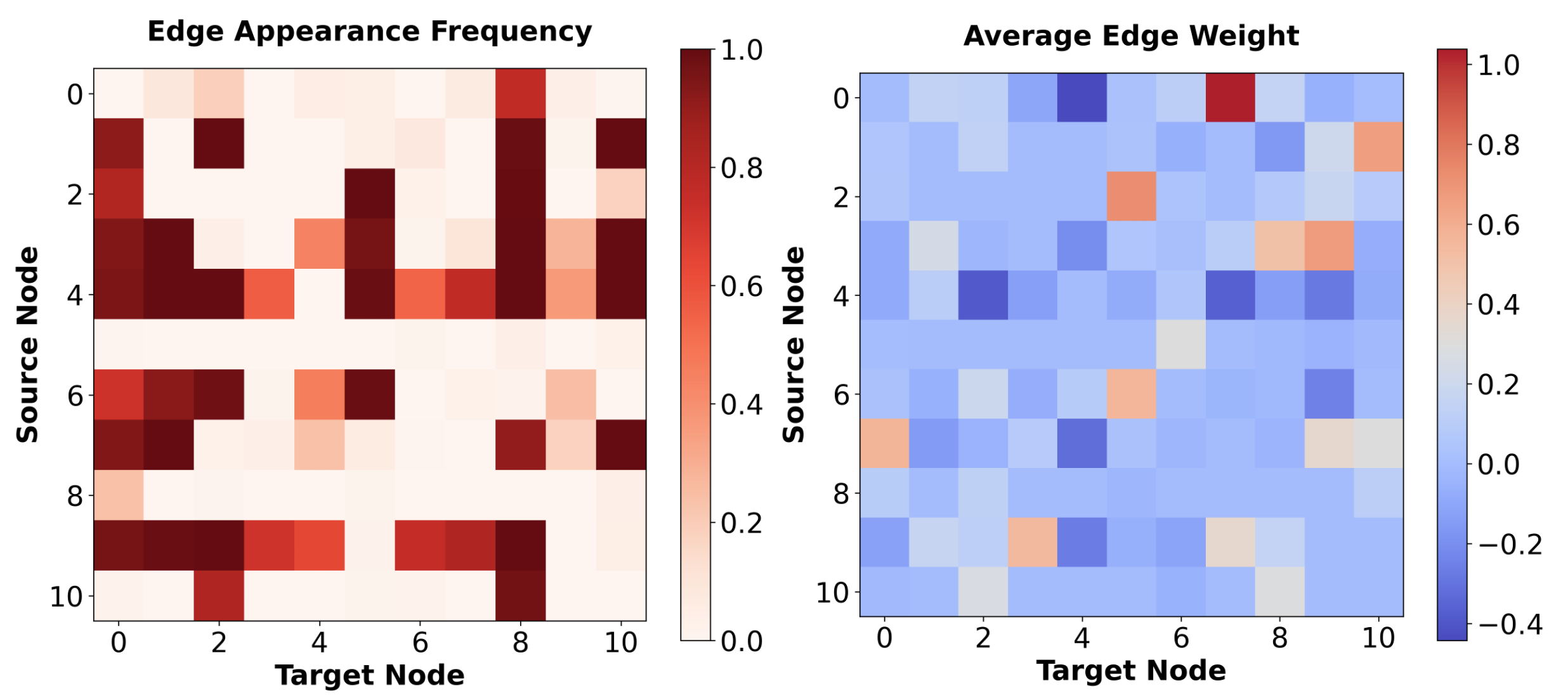}
    \caption{Cond-ERK: $\mathrm{CE}_{\mathrm{Erk}\to\mathrm{Akt}}>t_{\mathrm{ERK}}$}
    \label{fig:sachs_hm_cond_erk}
  \end{subfigure}\hfill
  \begin{subfigure}[t]{0.49\textwidth}
    \centering
    \includegraphics[width=\linewidth]{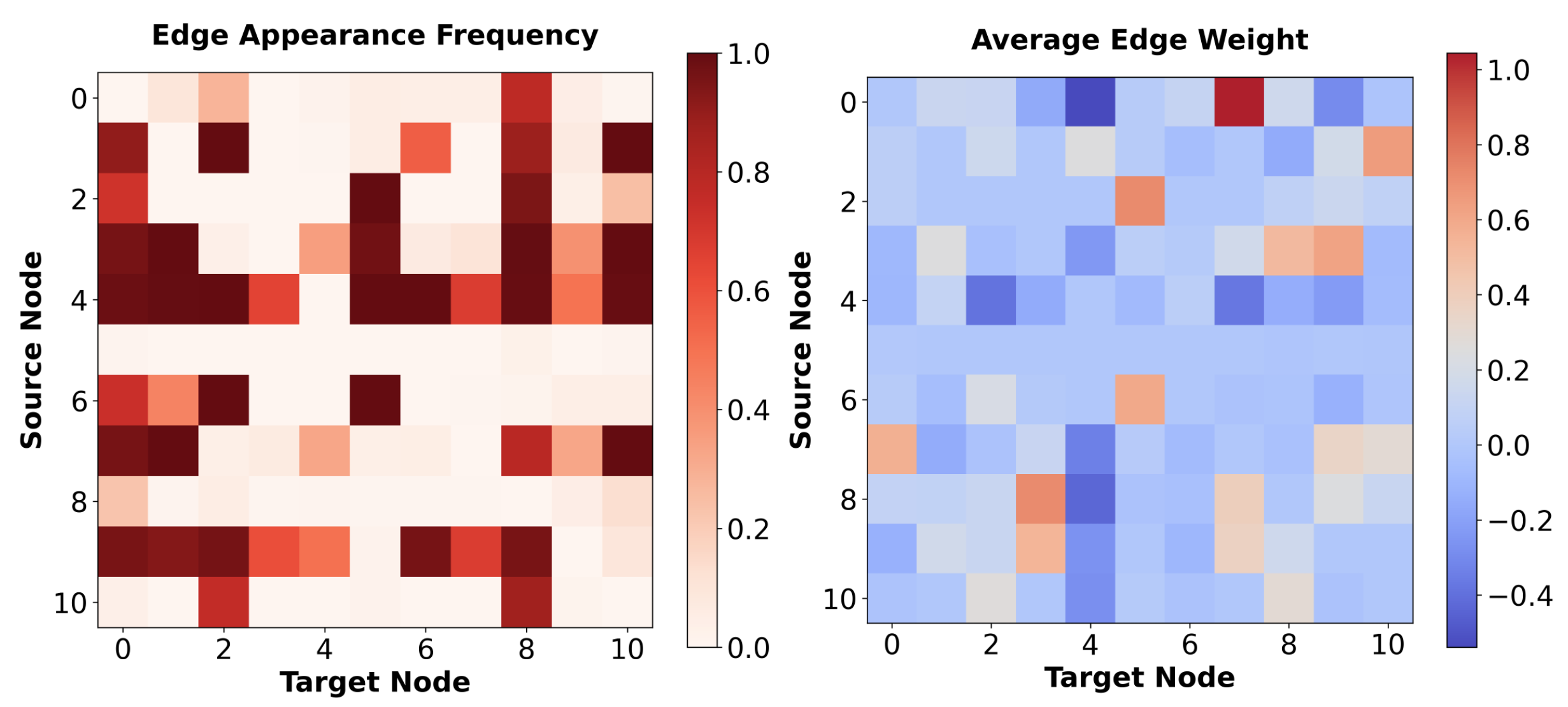}
    \caption{Cond-Joint: both constraints}
    \label{fig:sachs_hm_cond_joint}
  \end{subfigure}
  \caption{
 Heatmaps summarizing posterior graph features under four conditioning settings.
Each panel contains two heatmaps: \textbf{edge frequency} (probability an edge appears in sampled graphs; left) and \textbf{mean edge weight} (conditional mean of $B_{uv}$ given the edge is present; right).}
\label{fig:sachs_4_heatmaps}
\end{figure*}

\label{ssec:sachs_tables}
\begin{table*}[tbp]
\centering
\small
\caption{
Comparison across four conditioning settings for two target pairs. 
$\Pr(\mathrm{CE}>0)$ is the posterior mass with a directed influence (i.e., at least one directed path),
$\mathbb{E}[\mathrm{CE}\mid \mathrm{CE}>0]$ and quantiles are computed on the nonzero subset,
and \#paths is the median number of directed paths among samples with $\mathrm{CE}>0$.
We run PARNI-DAG single chain with $200,000$ iterations and collect $180000$ samples for the unconditioned case. For each of the other conditions, we collect $500$ survivor samples from 10 runs with $50$ samples for each run.}
\label{tab:sachs_4_effects}
\setlength{\tabcolsep}{4pt}
\begin{tabular}{lccc|ccc}
\toprule
& \multicolumn{3}{c}{\textbf{PIP3$\to$PIP2}} & \multicolumn{3}{c}{\textbf{Erk$\to$Akt}} \\
\cmidrule(lr){2-4}\cmidrule(lr){5-7}
\textbf{Condition}
& $\Pr(\mathrm{CE}>0)$ & $\mathbb{E}[\mathrm{CE}\mid \mathrm{CE}>0]$ & \#paths$_{\text{med}}$
& $\Pr(\mathrm{CE}>0)$ & $\mathbb{E}[\mathrm{CE}\mid \mathrm{CE}>0]$ & \#paths$_{\text{med}}$ \\
\midrule
Unconditioned
& 0.109 & \makecell{0.600 \\ {[0.543,\,0.681]}} & 4
& 0.085 & \makecell{0.569 \\ {[0.532,\,0.617]}} & 1 \\
\makecell[l]{Cond-PIP \\ ($\mathrm{CE}_{\mathrm{PIP}}>t_{\mathrm{PIP}}$)}
& 1.000 & \makecell{0.776 \\ {[0.771,\,0.782]}} & 2
& 0.498 & \makecell{0.596 \\ {[0.538,\,0.634]}} & 1 \\
\makecell[l]{Cond-ERK \\ ($\mathrm{CE}_{\mathrm{ERK}}>t_{\mathrm{ERK}}$)}
& 0.984 & \makecell{0.655 \\ {[0.562,\,0.710]}} & 10
& 1.000 & \makecell{0.673 \\ {[0.670,\,0.676]}} & 1 \\
\makecell[l]{Cond-Joint \\ ($\mathrm{CE}_{\mathrm{PIP}}>t'_{\mathrm{PIP}}\wedge \mathrm{CE}_{\mathrm{ERK}}>t'_{\mathrm{ERK}}$)}
& 1.000 & \makecell{0.747 \\ {[0.742,\,0.758]}} & 3
& 1.000 
& \makecell{0.656 \\ {[0.652,\,0.662]}} & 1 \\
\bottomrule
\end{tabular}
\end{table*}

\begin{table*}[tbp]
\centering
\small
\caption{Mechanism decomposition across conditions, using the two pathways most relevant for interpretation:
for PIP3$\to$PIP2 we use the direct path (PIP3$\to$PIP2) and the Plcg-mediated path (PIP3$\to$Plcg$\to$PIP2);
for Erk$\to$Akt we use the direct path (Erk$\to$Akt) and the Plcg-mediated path (Erk$\to$Plcg$\to$Akt).
For unconditioned samples, we select those with $\mathrm{CE}>0$, and calculate $\mathbb{E}[\mathrm{CE}\mid \mathrm{CE}>0]$.
The last column reports the frequency that the canonical mediated path edges are present (over all samples under the condition).
}
\label{tab:sachs_canonical_paths}
\begin{tabular}{llcccc}
\toprule
\textbf{Pair} & \textbf{Condition} &
$\mathbb{E}[\mathrm{CE}\mid \mathrm{CE}>0]$ &
Direct contrib &
Via-Plcg contrib &
$\Pr(\text{Via-Plcg path})$ \\
\midrule
\multirow{4}{*}{PIP3$\to$PIP2}
& Unconditioned & 0.600 & 0.560 (93.4\%) & 0.076 (12.7\%) & 0.064 \\
& Cond-PIP & 0.776 & 0.614 (79.2\%) & 0.161 (20.8\%) & 1.000 \\
& Cond-ERK & 0.655 & 0.559 (85.2\%) & 0.140 (21.4\%) & 0.975 \\
& Cond-Joint & 0.747 & 0.598 (79.9\%) & 0.154 (20.5\%) & 1.000 \\
\midrule
\multirow{4}{*}{Erk$\to$Akt}
& Unconditioned & 0.569 & 0.563 (98.9\%) & 0.000 (0.0\%) & 0.0004 \\
& Cond-PIP & 0.596 & 0.606 (101.6\%) & 0.001 (0.2\%) & 0.026 \\
& Cond-ERK & 0.673 & 0.668 (99.3\%) & 0.004 (0.6\%) & 0.177 \\
& Cond-Joint & 0.656 & 
0.650 (99.1\%) & 0.005 (0.8\%) & 0.244 \\
\bottomrule
\end{tabular}
\vspace{1mm}
\footnotesize

\textbf{Note:} Percentages are (path contribution)/(mean effect). They may exceed 100\% or be negative due to path cancellations.
\end{table*}

\subsection{Synthetic multi-effect heat maps}
\label{sec:synthetic_heatmaps_app}

\begin{figure}[t]
  \centering
  \begin{subfigure}[t]{\linewidth}
    \centering
    \includegraphics[width=\linewidth,trim=4 4 4 4,clip]{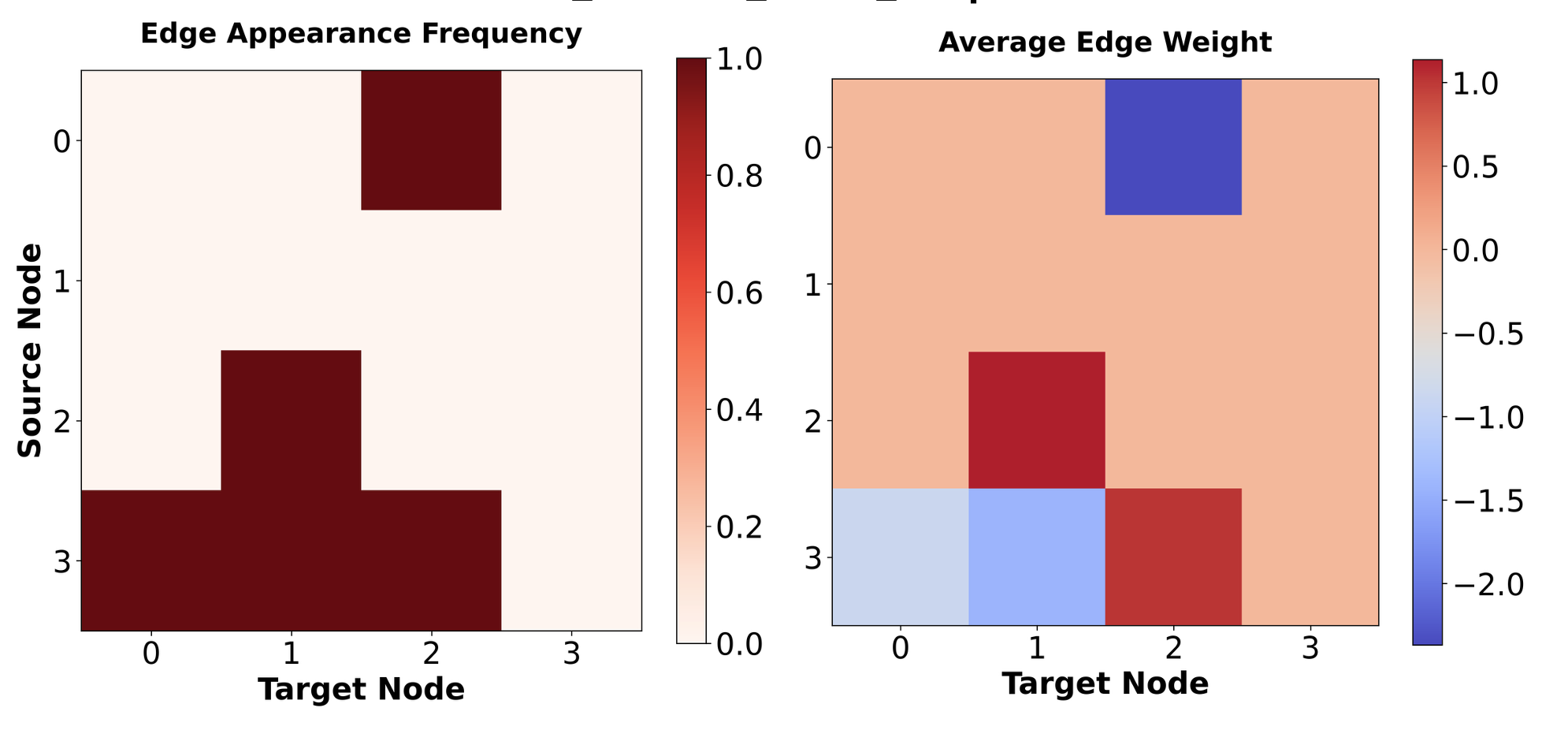}
    \caption{Ground-truth graph.}
  \end{subfigure}

  \vspace{1mm}
  \begin{subfigure}[t]{\linewidth}
    \centering
    \includegraphics[width=\linewidth,trim=4 4 4 4,clip]{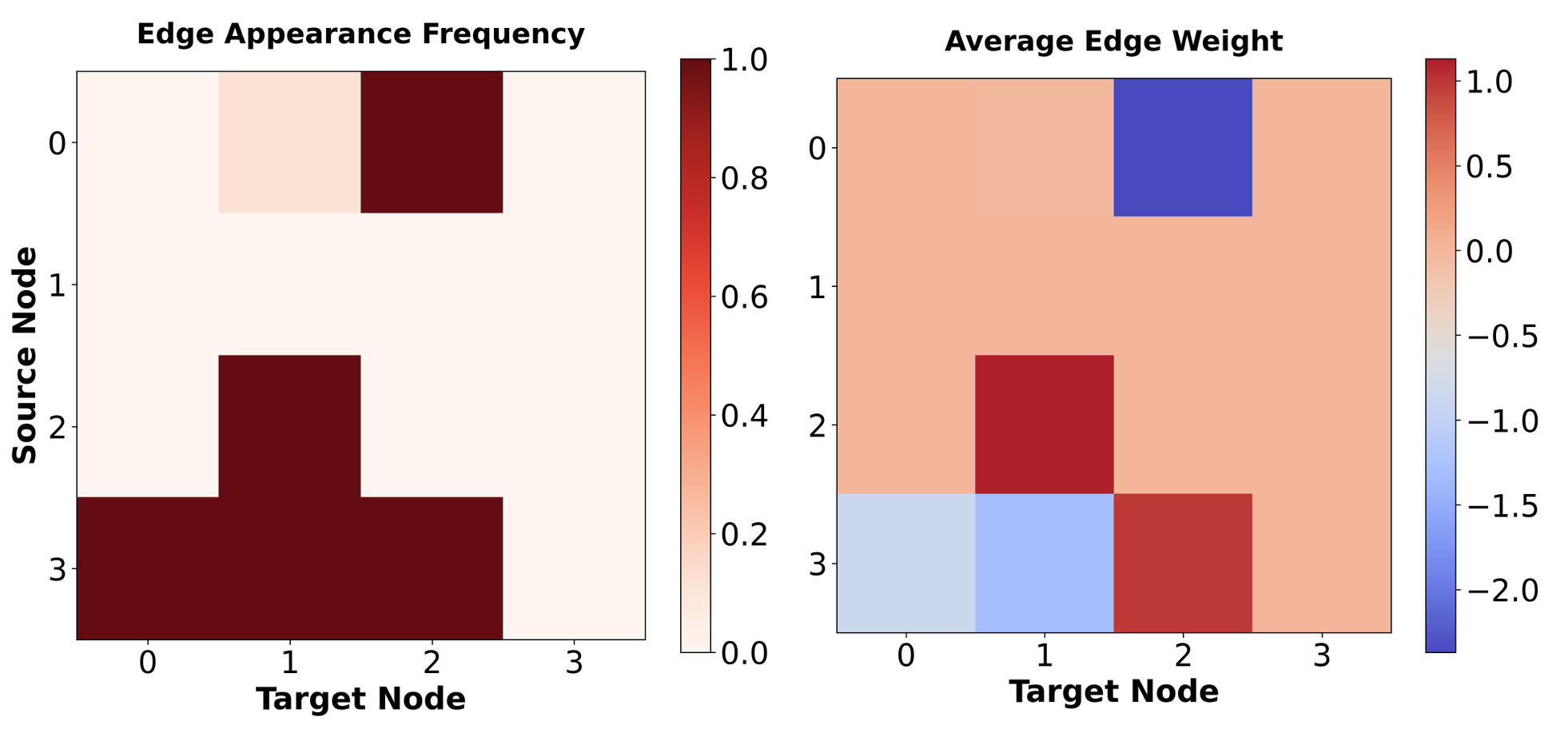}
    \caption{Conditional posterior under strong multi-effect constraints.}
  \end{subfigure}

  \vspace{1mm}
  \begin{subfigure}[t]{\linewidth}
    \centering
    \includegraphics[width=\linewidth,trim=4 4 4 4,clip]{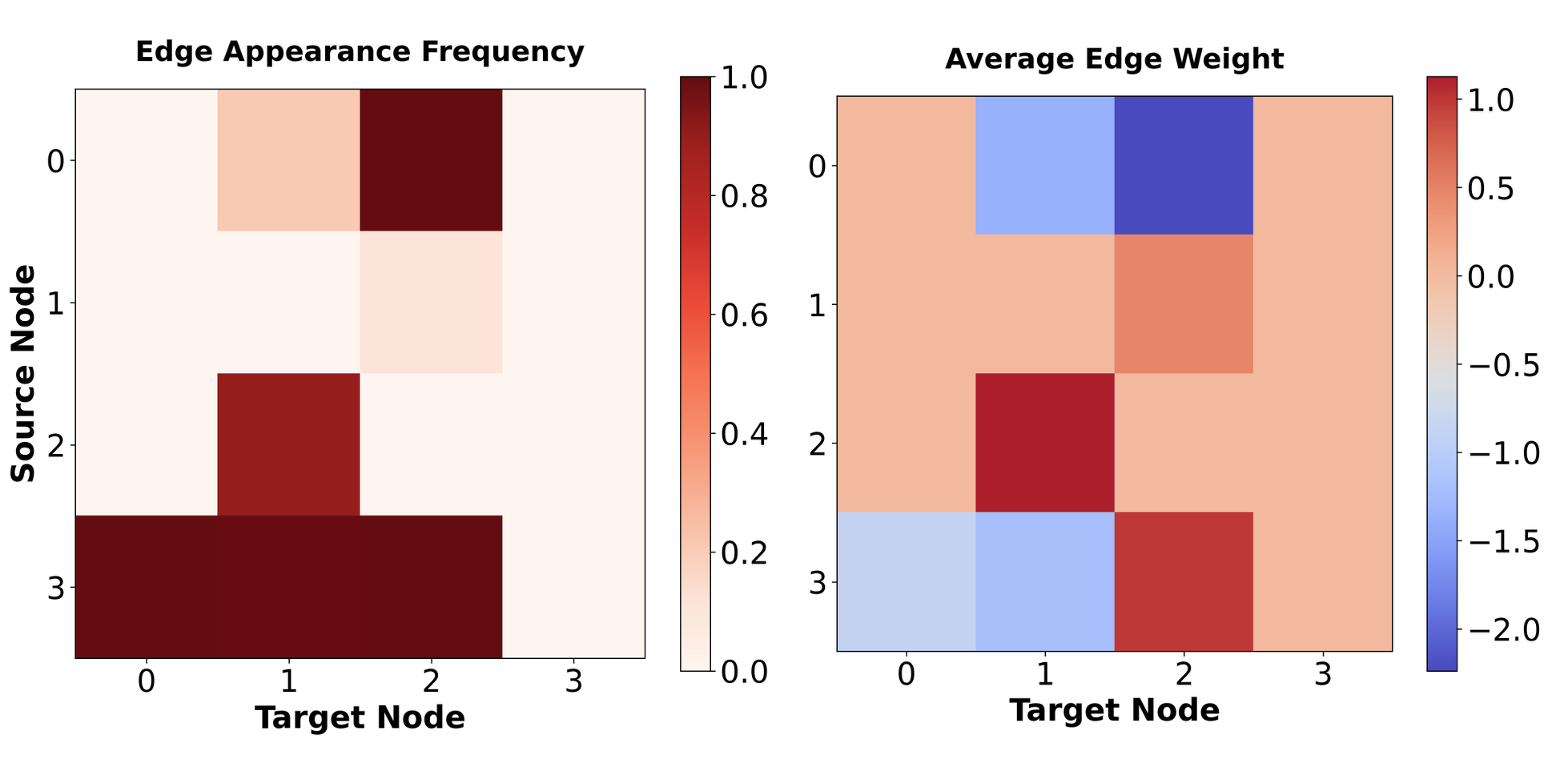}
    \caption{Conditional posterior under weak multi-effect constraints.}
  \end{subfigure}
  \caption{$d=4$ multi-effect conditioning results. Each panel summarizes edge
  frequencies and average edge weights. Stronger constraints concentrate
  posterior mass more tightly around the ground-truth mechanism.}
  \label{fig:d4_heatmaps}
\end{figure}

\begin{figure}[t]
  \centering
  \begin{subfigure}[t]{\linewidth}
    \centering
    \includegraphics[width=\linewidth,trim=4 4 4 4,clip]{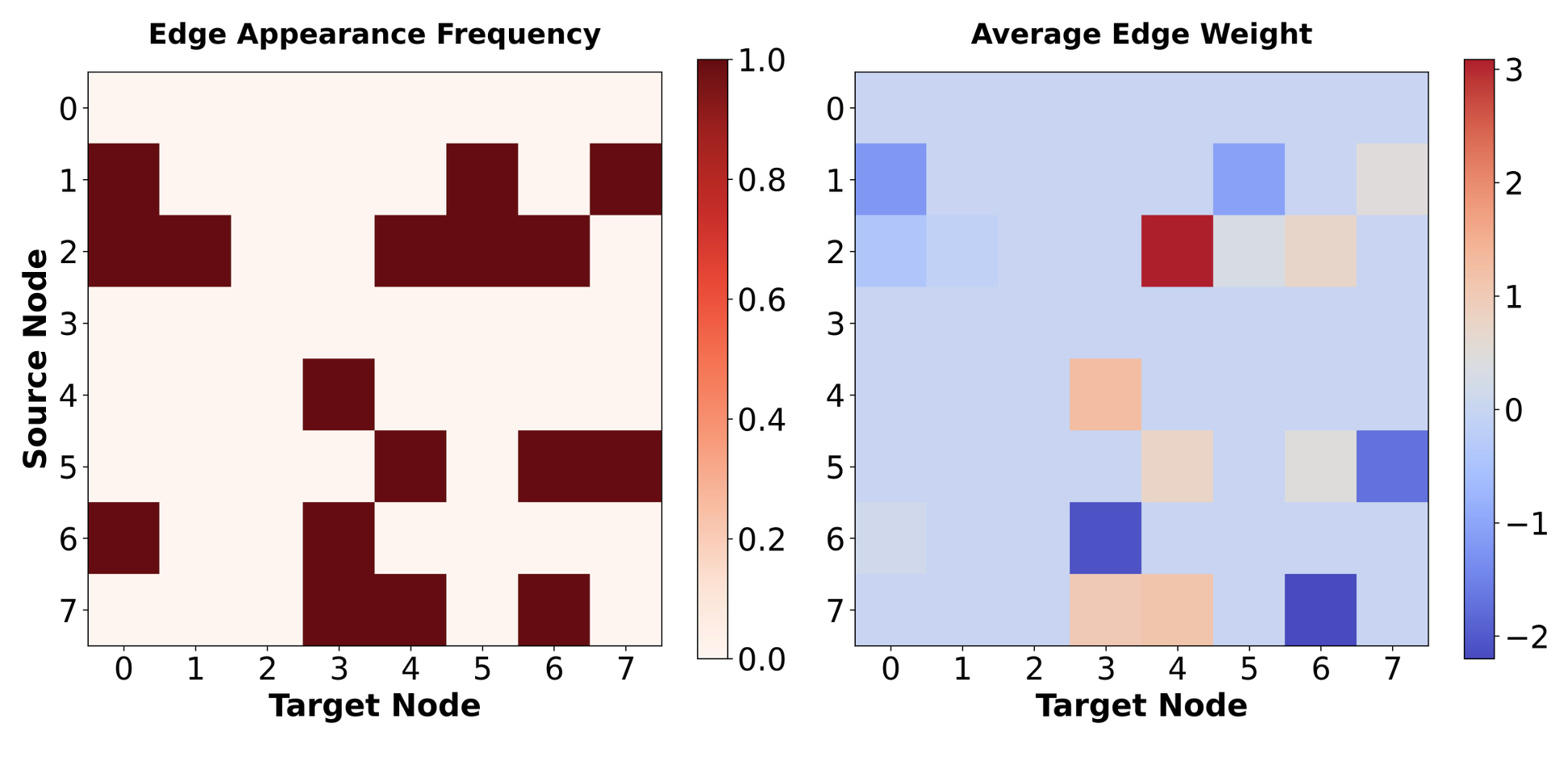}
    \caption{Ground-truth graph.}
  \end{subfigure}

  \vspace{1mm}
  \begin{subfigure}[t]{\linewidth}
    \centering
    \includegraphics[width=\linewidth,trim=4 4 4 4,clip]{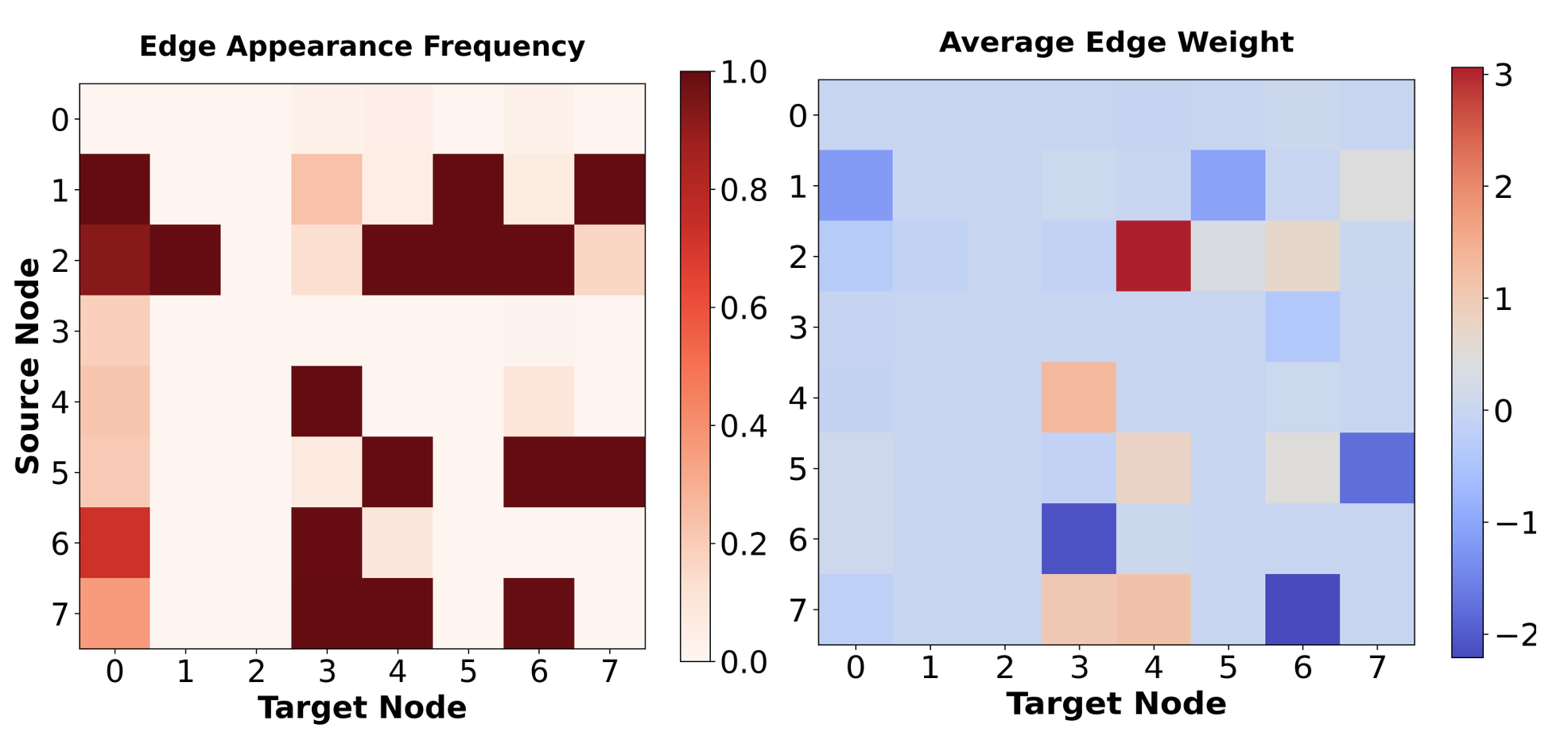}
    \caption{Conditional posterior under strong multi-effect constraints.}
  \end{subfigure}

  \vspace{1mm}
  \begin{subfigure}[t]{\linewidth}
    \centering
    \includegraphics[width=\linewidth,trim=4 4 4 4,clip]{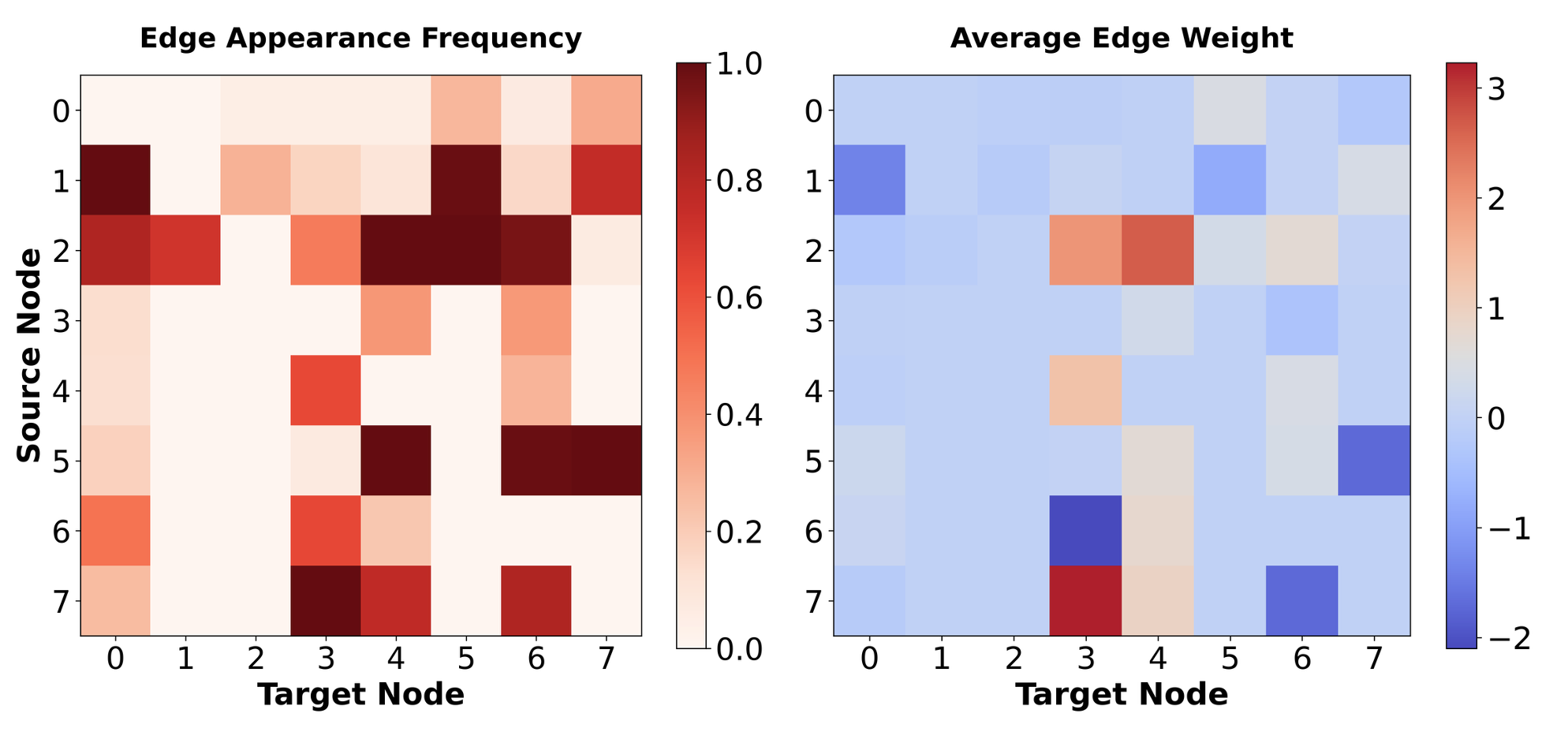}
    \caption{Conditional posterior under weak multi-effect constraints.}
  \end{subfigure}
  \caption{$d=8$ multi-effect conditioning results, with the same layout as
  Fig.~\ref{fig:d4_heatmaps}. The strong constraints produce sharper edge and
  weight summaries than the weak constraints.}
  \label{fig:d8_heatmaps}
\end{figure}

Figures~\ref{fig:d4_heatmaps} and~\ref{fig:d8_heatmaps} provide the detailed
heat-map visualizations for the multi-effect conditioning experiments discussed
in Sec.~\ref{sec:correctness_multi}. Each row displays the aggregated edge
frequency and average edge weight summaries for one posterior condition. These
figures are included in the appendix because they are visually dense; the main
text reports the more compact SHD summaries.

\subsection{Pseudo-code}
\label{ssec:pseudo-code}

Algorithm~\ref{alg:mls_framework} summarizes the adaptive multilevel-splitting
loop, and Algorithm~\ref{alg:mcmc_mutation} summarizes the level-truncated MCMC
mutation kernel. The notation matches Sec.~\ref{sec:method}: $h$ is the scalar
score, $\lambda_\star$ is the target level, and
$\mathcal A_{\lambda}=\{Z:h(Z)\ge\lambda\}$ is the corresponding score-level
event.

\begin{algorithm*}[tbp]
\caption{Adaptive Multilevel Splitting for Conditional Causal Discovery}
\label{alg:mls_framework}
\begin{algorithmic}[1]
\Require Data $\mathcal D$; posterior density $\pi(Z)=p(G,B\mid\mathcal D)$; score $h$; target level $\lambda_\star$; particle size $N$; survival fraction $\rho$; mutation steps $m$ per level; maximum number of levels $K_{\max}$; structure kernel $\mathsf{Kernel}$.
\Ensure Tail-probability estimate $\widehat p_\star$ and approximately constrained particles $\mathcal P$.

\State Initialize particles $\mathcal P=\{Z_n=(G_n,B_n)\}_{n=1}^{N}$ by drawing $G_n\sim p(G)$ and $B_n\sim p(B\mid G_n,\mathcal D)$.
\State Set $\lambda_0\gets -\infty$, $\widehat p_\star\gets 1$, and $k\gets 0$.
\State $\mathcal P\gets \Call{MCMCMutation}{\mathcal P,\mathcal D,h,\lambda_0,m,\mathsf{Kernel}}$.
\While{$k<K_{\max}$}
    \State Compute scores $s_n\gets h(Z_n)$ for all $Z_n\in\mathcal P$.
    \State $\widetilde\lambda_{k+1}\gets Q_{1-\rho}(\{s_n\}_{n=1}^{N})$.
    \State $\lambda_{k+1}\gets \min\{\widetilde\lambda_{k+1},\lambda_\star\}$.
    \State $\mathcal S\gets \{Z_n\in\mathcal P:h(Z_n)\ge\lambda_{k+1}\}$.
    \State $\widehat\beta_{k+1}\gets |\mathcal S|/N$.
    \State $\widehat p_\star\gets \widehat p_\star\,\widehat\beta_{k+1}$.
    \If{$|\mathcal S|=0$}
        \State \Return $(0,\emptyset)$.
    \EndIf
    \State Resample $N$ particles from $\mathcal S$ with replacement to form $\mathcal P$.
    \State $\mathcal P\gets \Call{MCMCMutation}{\mathcal P,\mathcal D,h,\lambda_{k+1},m,\mathsf{Kernel}}$.
    \If{$\lambda_{k+1}=\lambda_\star$}
        \State \Return $(\widehat p_\star,\mathcal P)$.
    \EndIf
    \State $k\gets k+1$.
\EndWhile
\State \Return $(\widehat p_\star,\mathcal P)$.
\end{algorithmic}
\end{algorithm*}

\begin{algorithm*}[tbp]
\caption{Level-Truncated MCMC Mutation over Graph--Weight States}
\label{alg:mcmc_mutation}
\begin{algorithmic}[1]
\Require Population $\mathcal P$; data $\mathcal D$; score $h$; level $\lambda$; mutation steps $m$; structure kernel $\mathsf{Kernel}$; structure-move probability $p_{\mathrm{struct}}$.
\Ensure Mutated population $\mathcal P'$.

\State $\mathcal P'\gets\emptyset$.
\ForAll{$Z=(G,B)\in\mathcal P$}
  \For{$r=1$ \textbf{to} $m$}
    \State Draw $u\sim\mathrm{Uniform}(0,1)$.
    \If{$u<p_{\mathrm{struct}}$}
        \State Propose a graph $G'\sim q_G(\cdot\mid G)$ using $\mathsf{Kernel}$.
        \State Refresh affected coefficient blocks from $p(B\mid G',\mathcal D)$ and copy unchanged blocks to obtain $B'$.
    \Else
        \State Set $G'\gets G$ and refresh one coefficient block from $p(B\mid G,\mathcal D)$ to obtain $B'$.
    \EndIf
    \State Set $Z'\gets(G',B')$ and let $q(Z'\mid Z)$ denote the complete proposal density.
    \If{$h(Z')<\lambda$}
        \State Reject $Z'$ and continue.
    \Else
        \State $\alpha\gets \min\!\left\{1,\dfrac{\pi(Z')q(Z\mid Z')}{\pi(Z)q(Z'\mid Z)}\right\}$.
        \State Draw $a\sim\mathrm{Uniform}(0,1)$.
        \If{$a\le\alpha$}
            \State $Z\gets Z'$.
        \EndIf
    \EndIf
  \EndFor
  \State $\mathcal P'\gets\mathcal P'\cup\{Z\}$.
\EndFor
\State \Return $\mathcal P'$.
\end{algorithmic}
\end{algorithm*}

\end{document}